\documentclass[10pt,twocolumn,letterpaper]{article}

\usepackage{iccv}              

\usepackage{url}
\usepackage{multicol} 
\usepackage{array} 
\usepackage{threeparttable}
\usepackage{cutwin}
\usepackage{subcaption}
\usepackage{graphicx} 
\usepackage{wrapfig}
\usepackage{xcolor}
\usepackage{colortbl}
\usepackage[none]{hyphenat} 

\usepackage{booktabs}
\usepackage{multirow}
\usepackage{dsfont}
\usepackage{graphicx}	
\usepackage{amsmath}	
\usepackage{amssymb}	
\usepackage{booktabs}
\usepackage{times}
\usepackage{microtype}   

\usepackage{epsfig}
\usepackage{caption}
\usepackage{float}
\usepackage{placeins}
\usepackage{color, colortbl}
\usepackage{stfloats}
\usepackage{enumitem}
\usepackage{tabularx}
\usepackage{xstring}
\usepackage{xspace}
\usepackage{url}
\usepackage{subcaption}
\usepackage{xcolor}
\usepackage[hang,flushmargin]{footmisc}
\usepackage[ruled]{algorithm2e}
\usepackage{multirow}

\definecolor{iccvblue}{rgb}{0.21,0.49,0.74}
\usepackage[pagebackref,breaklinks,colorlinks,allcolors=iccvblue]{hyperref}
\def\ourname{LocalDyGS}
\def\paperTitle{\ourname: Multi-view Global Dynamic Scene Modeling via \\ Adaptive Local Implicit Feature Decoupling}

\def\paperID{6878} 
\def\confName{ICCV}
\def\confYear{2025}

\title{\paperTitle}

\author{%
  Jiahao Wu$^{1,2}$, \quad Rui Peng$^{1,2}$, \quad Jianbo Jiao$^3$, \quad Jiayu Yang$^2$,   Luyang Tang$^{1,2}$ \\ 
  \quad Kaiqiang Xiong$^{1,2}$, \quad Jie Liang$^1$ \quad Jinbo Yan$^1$, \quad  Runling Liu$^1$ \quad Ronggang Wang$^{1,2}\dagger$ \\
  $^1$ Guangdong Provincial Key Laboratory of Ultra High Definition Immersive Media Technology \\
  Shenzhen Graduate School, Peking University\\
  $^2$ Pengcheng Lab ~~~ $^3$ University of Birmingham
}

\begin{document}

\twocolumn[{%
\renewcommand\twocolumn[1][]{#1}%
\maketitle
\begin{center}
\centering
\includegraphics[width=\textwidth, trim=0cm 9.5cm 4cm 1.4cm]{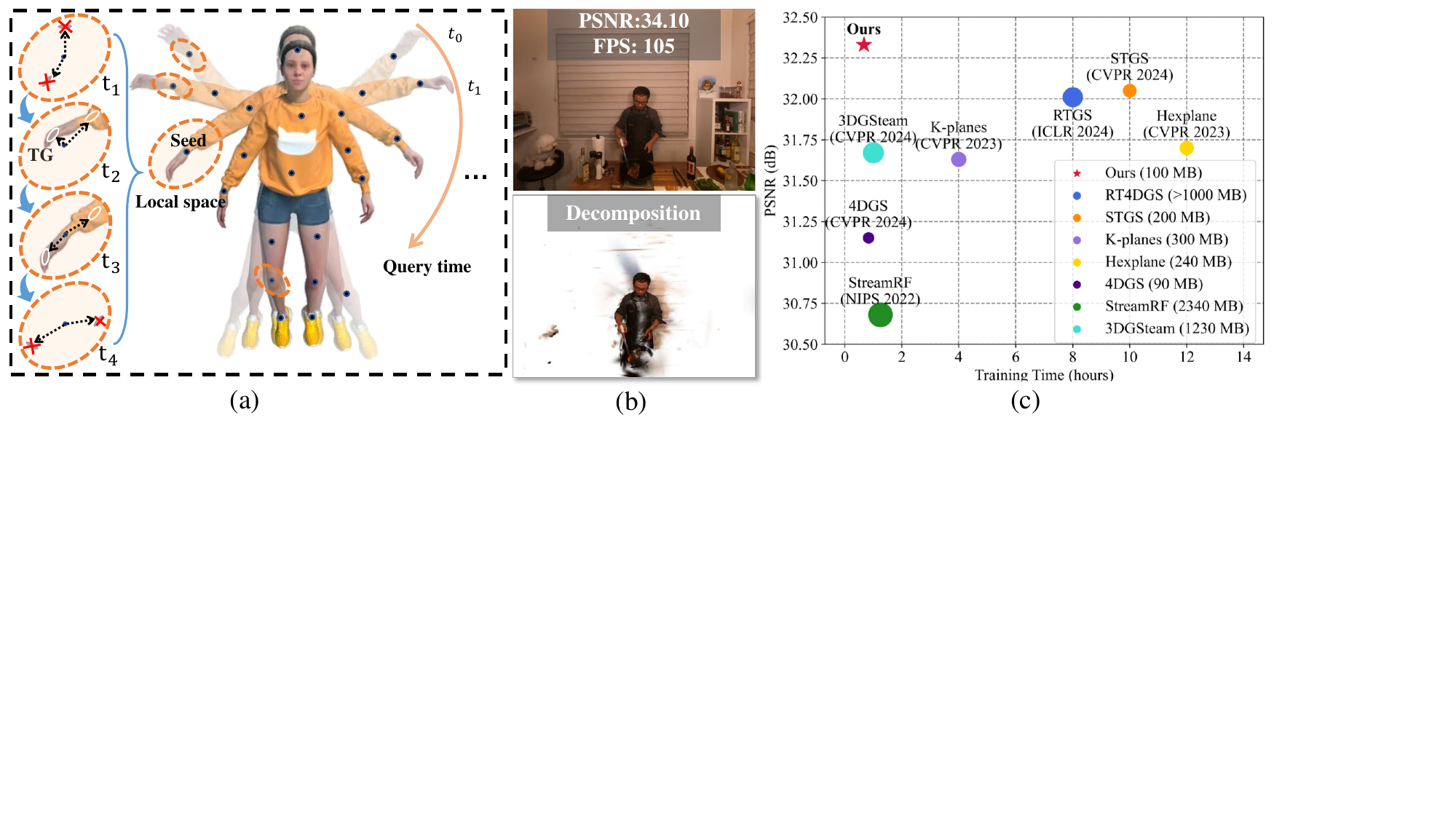}
\captionof{figure}{(a) shows our foundational idea: Decomposing a globally complex dynamic scene into a series of streamlined local spaces. The Temporal Gaussian (TG) activates only when the arm enters the local space, generating varying TGs to represent the motion of the arm, and deactivates once the arm exits. (b) displays our high-quality rendering results and the accuracy of dynamic-static decoupling, while (c) demonstrates the superior performance of our method compared to other approaches on the N3DV~\cite{li2022neural} dataset. 
}
\label{fig:teaserfig}
\end{center}%
}]

\renewcommand{\thefootnote}{\fnsymbol{footnote}}\footnotetext[2]{Corresponding author.}

\begin{abstract}

\vspace{-3mm}

Due to the complex and highly dynamic motions in the real world, synthesizing dynamic videos from multi-view inputs for arbitrary viewpoints is challenging. Previous works based on neural radiance field or 3D Gaussian splatting are limited to modeling fine-scale motion, greatly restricting their application. 
In this paper, we introduce \ourname, which consists of two parts to adapt our method to both large-scale and fine-scale motion scenes:
1) We decompose a complex dynamic scene into streamlined local spaces defined by seeds, enabling global modeling by capturing motion within each local space.
2) We decouple static and dynamic features for local space motion modeling. A static feature shared across time steps captures static information, while a dynamic residual field provides time-specific features. These are combined and decoded to generate Temporal Gaussians, modeling motion within each local space.
As a result, we propose a novel dynamic scene reconstruction framework to model highly dynamic real-world scenes more realistically. Our method not only demonstrates competitive performance on various fine-scale datasets compared to state-of-the-art (SOTA) methods, but also represents the first attempt to model larger and more complex highly dynamic scenes.  Project page: \url{https://wujh2001.github.io/LocalDyGS/}.

\end{abstract}

\section{Introduction}
\label{sec:intro}

Multi-view dynamic scene reconstruction is a crucial and challenging problem with a wide range of applications, such as free-viewpoint control for sports events and movies, AR, VR, and gaming. 
For monocular dynamic scene reconstruction, due to the lack of accurate geometric information in monocular input, it is often limited to reconstructing relatively simple scenes \cite{park2021hypernerf,pumarola2021d,yan2023nerf} and struggles to model highly dynamic and complex scenes, thus failing to provide users with a more immersive visual experience. To model the complex and dynamic real-world scenes with high quality, a widely adopted solution is to use multi-view synchronized videos to provide dense spatiotemporal supervision \cite{wang2022mixed,li2022neural,bansal20204d,lombardi2019neural,zitnick2004high}.

Many researchers have explored multi-view dynamic scene reconstruction from different perspectives to enhance visual quality. For example, 3DGStream \cite{sun20243dgstream} utilizes a Neural Transformation Cache (NTC) to model each frame individually, enabling streaming dynamic scene reconstruction. SpaceTimeGS \cite{li2024spacetime} employs polynomials to control the motion trajectories and opacity of Gaussian points, thereby representing the entire dynamic scene. More recently, Swift4D \cite{wuswift4d} leverages pixel variance as a supervisory signal to decouple static and dynamic Gaussian points. Meanwhile, they validate their approach using a basketball court dataset \cite{VRU} with larger motion scales and a more complex environment. Despite significant progress, challenges remain: 1) flickering and blurring issues with large-scale complex motion datasets, and 2) high training time and storage requirements.

\begin{figure}
   \centering
   \includegraphics[width=0.8\textwidth, trim=0cm 10.5cm -1cm 0cm]{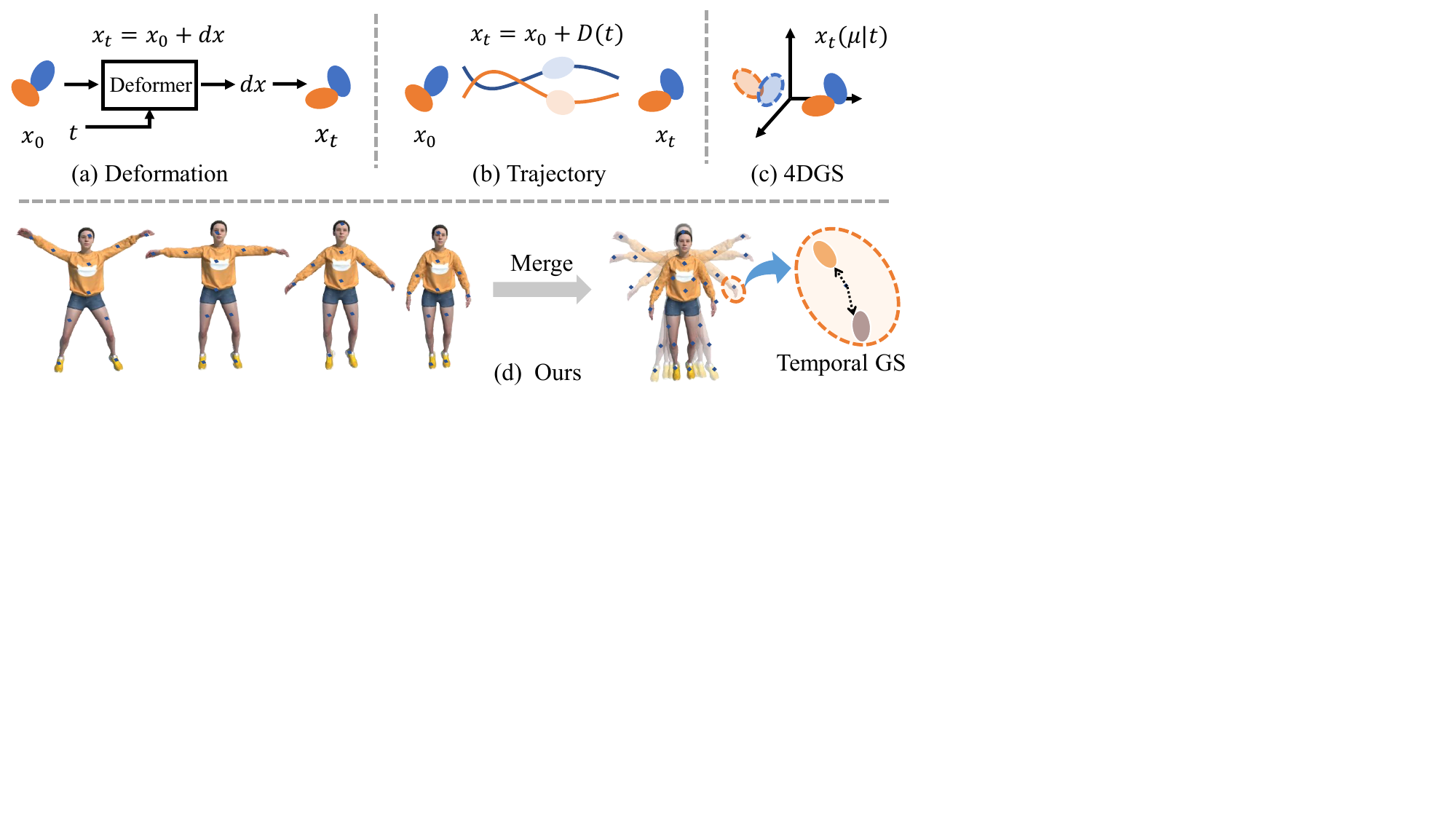}
   \caption{Overview of existing dynamic methods.}
   \label{fig: method_compare}
   \vspace{-10pt}
\end{figure}


Therefore, we propose \textit{\ourname}, a multi-view dynamic method adaptable to both fine and large-scale motion, comprising two-fold: 1) decomposing the global space into local spaces, and 2) generating Temporal Gaussians to model motion within each local space ( \textit{Local space} is defined as the space surrounding a seed ). 
Specifically, our method no longer explicitly models the longtime motion of each Gaussian point. Instead, as shown in Fig. \ref{fig:teaserfig} (a), we use seeds to decompose the complex 3D space into an array of independent local spaces. 
Local motion modeling is then achieved by generating Temporal Gaussian within each local space, enabling global dynamic scene representation. 
When seeds cover all regions where a moving object appears, this local space motion modeling approach has the potential to handle large-scale dynamic scenes.
To ensure complete coverage, we use a fused Structure from Motion (SfM) \cite{schonberger2016structure} point cloud from multiple frames, positioning seed points across all areas with moving objects.

For specific details on local space motion modeling, we assign a learnable static feature shared across all time steps to represent time-invariant static information. As for the dynamic information within each local space, we also construct a dynamic residual field that provides unique dynamic residual features at each time step. With the static feature as a base, dynamic features at each time step generate distinct Temporal Gaussians to model motion over time. This decoupled design helps reduce the load on the dynamic residual field. 
Next, an adaptive weight field is designed to balance static and dynamic residual features. These features are combined through a weighted linear sum and decoded by a dedicated multilayer perceptron (MLP) to produce the corresponding Temporal Gaussian, capturing motion within the local space. 
Finally, we propose an adaptive error-based seed growth strategy to alleviate incomplete coverage in the initial point cloud, thereby improving the model's robustness to the SfM \cite{schonberger2016structure} initialized point cloud.  In summary, our contributions can be outlined as follows:

\begin{itemize}
\item We propose to decompose the 3D space into seed-based local spaces, enabling global dynamic scene modeling with the capacity for multi-scale motion.
\item We propose decomposing scene features into static and dynamic components to simplify local dynamic modeling and enhance rendering quality.
\item We designed a unique Adaptive Seed Growing strategy to address the issue of incomplete coverage of dynamic scenes by the initial point cloud. 
\item We are the first to extend dynamic reconstruction to a large scale, and extensive experiments validate our superior performance across various metrics.
\end{itemize}

\begin{figure*}
   \centering
   \includegraphics[width=\textwidth, trim=0cm 9cm 0.5cm 0cm]{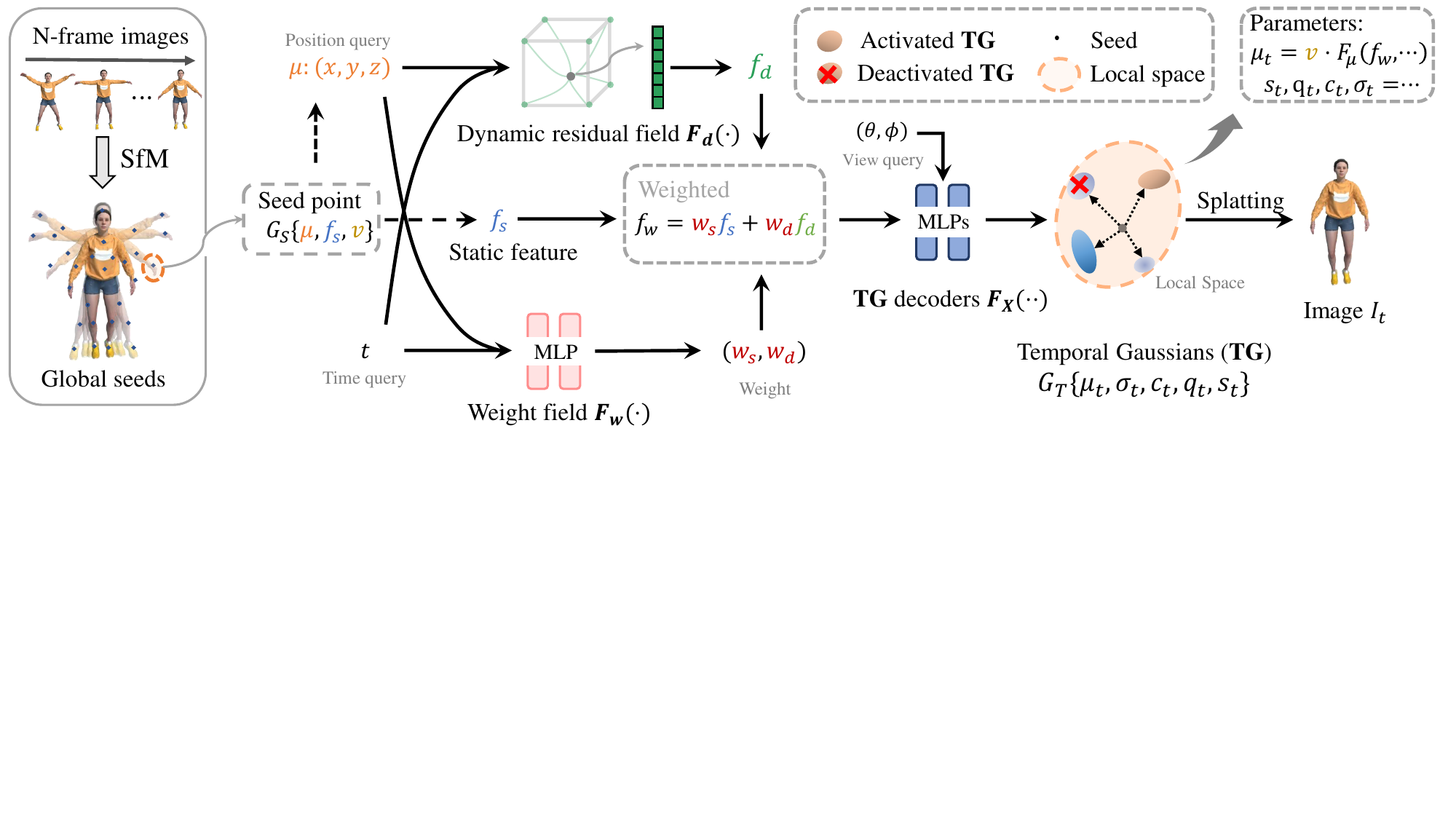}
   \caption{ \textbf{Overview of \ourname.} 
    We sample \( N \) frames across the time domain to extract the SfM \cite{schonberger2016structure} point cloud, using it to initialize seeds and local spaces, with each seed assigned two learnable parameters: a static feature \( f_s \) shared across all time steps, and a scale \( v \) defining the local space range.
   Additionally, we construct a global dynamic residual field and a weighting field to provide temporal information for the local space. The two are combined through weighted linear summation to obtain the weighted feature \( f_w \), which is then passed through a dedicated Temporal  Gaussian (TG) decoder to predict parameters such as mean and color of the Temporal  Gaussians. Finally, we perform a deactivation operation to remove Temporal Gaussians that do not belong to the query time \( t \) for rasterization. 
   \textit{We use the 'jumping' sequence from the  D-NeRF monocular dataset for demonstration, but our method is based on multi-view reconstruction.}
   }
   \label{fig: pipeline}
\end{figure*}

\section{Related Work}
\label{sec:related}

\subsection{Novel View Synthesis for Static Scenes.} 
Synthesizing novel views for static scenes is a classical and well-studied problem. Previous years, NeRF has emerged as a groundbreaking work in novel view synthesis, inspiring a series of new view synthesis approaches aimed at improving training speed and rendering quality \cite{barron2021mip,barron2022mip,hu2023tri,barron2023zip,chen2022tensorf,fridovich2022plenoxels}, surface reconstruction \cite{wang2021neus,wang2023neus2}, autonomous driving \cite{wu2023mars,wu2024dynamic}, SLAM \cite{zhu2022nice} etc.. Recently, 3DGS \cite{kerbl20233d} has garnered significant attention in the community for its rapid model training and real-time inference, achieving SOTA visual quality. Many advanced works have emerged, focusing on surface reconstruction~\cite{huang20242d,zhang2024rade,radl2024stopthepop,yu2024gaussian}, few-shot~\cite{zhu2025fsgs,fan2024instantsplat,yu2024viewcrafter} and pose-free methods~\cite{Fu_2024_CVPR,bortolon20246dgs},  HDR \cite{wu2024hdrgs,xiao2025multi}. In particular, recent works~\cite{lu2024scaffold,ren2024octree} suggest that world space is sparse and can be represented using a set of structural points to represent a class of points to achieve more compact 3D scene representation \cite{lu2024scaffold,wan2024superpoint,huang2024sc}.

\subsection{Novel View Synthesis for Dynamic Scenes.} 
Synthesizing novel views for dynamic scenes is a more challenging and applicable problem. 
A variety of NeRF-based dynamic scene methods, such as deformation field  \cite{park2021nerfies,gao2021dynamic,pumarola2021d}, scene flow  \cite{li2021neural}, and multi-plane  \cite{cao2023hexplane,fridovich2023k}, have been proposed. Among these, \cite{song2023nerfplayer} and \cite{wang2023mixed} specifically decouple dynamic and static elements to achieve higher rendering quality. 
A more related topic to our work, 3DGS-based dynamic methods, has emerged in recent literature and can be roughly categorized into three types:
As shown in Fig. \ref{fig: method_compare} (a),  deformation field methods \cite{wu20244d,yang2024deformable,huang2024sc,bae2024per,wan2024superpoint,zhao2024gaussianprediction, yan2025instant}, represented by \cite{wu20244d}, which map Gaussian points in a canonical field to a deformation field to represent dynamic scenes at each timestamp. 
As shown in Fig. \ref{fig: method_compare} (b), trajectory tracking-based solutions \cite{kratimenos2023dynmf,lin2024gaussian} typically use polynomials or Fourier series to represent the motion trajectory of each Gaussian. 
As shown in Fig. \ref{fig: method_compare} (c), methods that extend 3DGS to 4DGS \cite{yang2023real,duan20244d} require a large number of Gaussian points for fitting, resulting in high storage requirements and slower training speeds.

 \textbf{Monocular and multi-view dynamic scene.} Although many recent monocular dynamic reconstruction works \cite{yang2024deformable,park2021hypernerf,huang2024sc,som2024} have advanced the field, relying solely on monocular video as input remains challenging for reconstructing complex real-world scenes. Current methods are still limited to synthetic datasets \cite{pumarola2021d} or simple motion scenarios \cite{park2021hypernerf,yan2023nerf}. In contrast, for complex real-world reconstruction, leveraging multi-view synchronized videos to provide dense spatiotemporal supervision appears to be more promising.  
Dynerf \cite{li2022neural}, 3DGStream \cite{sun20243dgstream}, and SpaceTimeGS \cite{li2024spacetime} etc. have explored multi-view dynamic scenes, demonstrating the potential of free-viewpoint outputs from multi-view inputs. However, as shown in our experiments, they suffer from blurring and flickering in real-world scenes with complex, large-scale motion, limiting their applicability. To address this, we propose \textit{\ourname}, which handles larger-scale and fine-scale motion scenes with a more compact structure, faster training speed, and higher-quality rendering.

\section{Method}
\label{sec:method}
In this section, we introduce \ourname~, which consists of two main components: 1) decomposing the global space into local spaces and 2) generating Temporal Gaussians to model motion within each local space, as shown in Fig. \ref{fig: pipeline}.  
In the following subsections, we first describe the initialization of our seeds and local spaces. Next, we introduce the spatio-temporal fields, which equip each local space with essential temporal information for dynamic modeling. Finally, we explain the densification process and the training of our method.

\subsection{3DGS and ScaffoldGS Preliminary}
\label{subsec: Preliminary}
As an emerging popular technique for novel view synthesis, 3DGS \cite{kerbl20233d} uses 3D Gaussians as rendering primitives. Each primitive is defined as \( G\{\mu, q, s, \sigma, c\} \), where the parameters represent mean (\(\mu\)), rotation (\(q\)), scaling (\(s\)), opacity (\(\sigma\)), and color (\(c\)), respectively. A 3D Gaussian point \( G(x) \) can be mathematically defined as:
\begin{equation} 
 G(x) = e^{-\frac{1}{2}(x-\mu)^T\Sigma^{-1}(x-\mu)} 
\end{equation} 
where \(\Sigma\) is the covariance of the 3D Gaussian, typically represented by \(q\) and \(s\). During the rendering stage, as described in \cite{zwicker2002ewa}, the 3D Gaussian is projected into a 2D Gaussian \( G'(x) \). Then, the rasterizer sorts the 2D Gaussians and applies \(\alpha\)-blending.
\begin{equation}
\resizebox{.9\linewidth}{!}{$
    C(p)=\sum_{i\in K}c_i\alpha_i(p)\prod_{j=1}^{i-1}(1-\alpha_j(p)),~~\alpha_i(p)=\sigma_iG_i'(p).
$}
\end{equation}

Here, \( p \) represents the position of the pixel, and \( K \) denotes the number of 2D Gaussians intersecting with the queried pixel. Finally, end-to-end training can be achieved through supervised views.

A work closely related to ours is the static reconstruction method ScaffoldGS \cite{lu2024scaffold}, in which the scene is represented using anchors. Each anchor is associated with the following attributes: a mean position $\mu_a \in \mathbb{R}^{3}$, a static feature vector $f_a \in \mathbb{R}^{32}$, a scale factor $l_a \in \mathbb{R}^{3}$, and offsets $O_a \in \mathbb{R}^{k \times 3}$ corresponding to $k$ Gaussian points. The positions of neural Gaussians are calculated as:
\begin{equation}
    \{\mu_0,...,\mu_{k-1}\}=\mu_a+\{\mathcal{O}_0,...\mathcal{O}_{k-1}\}\cdot l_a.
\end{equation}
In addition, the other Gaussian parameters are also decoded using MLPs. To distinguish from the Scaffold \textit{anchor}, we use the \textit{Seed} to refer to the anchor in our dynamic method.

\subsection{Global Seeds Initialization}
\label{subsec: seed init}

In our framework, as shown in Fig. \ref{fig: pipeline}, \ourname~fuses SfM point clouds from \( N \) frames across the time domain to initialize seed positions \( \mu \), providing prior knowledge of where dynamic objects appear.
One of our core ideas is that each sparse seed point models the temporal dynamics of only its surrounding 3D scene (referred to as local space), rather than performing long-term motion tracking as in previous methods \cite{sun20243dgstream, li2024spacetime}. This means we allow a moving object to be represented by multiple seeds. As shown in Fig. \ref{fig: pipeline}, the moving arm is modeled by a series of different seeds. This significantly reduces the complexity of motion modeling.

For local space modeling, since static information occupies a large portion of the scene and varies significantly across each local space, we assign each local space an independently optimized static feature \( f_s \in \mathbb{R}^{64} \) to more accurately capture the static information. This feature is shared across all time steps and initialized to 0.
Additionally, we assign each local space a scale parameter \( v \in \mathbb{R}^{3} \) to control the spatial range of its influence. It is initialized as the average distance between the three nearest seed points. In areas with sparser seed points, the local space for each seed becomes larger. Finally, a seed in local space can be defined as \( G_\mathcal{S}\{\mu, f_s, v\} \). 
\begin{itemize}
    \item the position of seed (global parameter)
    \item static feature of local space (shared across all time steps)
    \item the scale of local space 
\end{itemize}

\subsection{ Feature-Decoupled Spatio-Temporal Fields}
\label{subsec:decomposed}

At first, we attempted to model the entire scene using a single spatio-temporal structure (without static features), following previous methods \cite{wu2024dynamic}. However, we found that this approach causes blurring in both dynamic and static regions, as shown in Fig. \ref{fig: wo static}.
We speculate that a single model struggles to store such vast scene information. Inspired by the Deformable-based method \cite{wu2024dynamic,wuswift4d,yang2024deformable}, they use the canonical field as a base and introduce a time-aware deformation field to reconstruct dynamic scenes.  
At a more fundamental feature level, we decouple scene information into static and dynamic residual features. Specifically, for each local space, we use independent static features \( f_s \in  \mathbb{R}^{64}\) as the foundation, capturing most of the local space’s static information, while a shared dynamic residual field $F_d$ encodes temporal variations for each local space, enabling dynamic scene reconstruction.

Since motion often exhibits local similarity, we need a compact, adaptive structure to deliver dynamic residual features while preserving locality to ensure neighboring seeds share similar features.
Inspired by~\cite{muller2022instant}, we construct the dynamic residual field by combining multi-resolution four-dimensional hash encoding with a shallow, fully-fused MLP. 
Specifically, each voxel grid node at different resolutions is mapped to a hash table storing \( d \)-dimensional learnable feature vectors. Given a seed point and query time \( (\mu, t) \in \mathbb{R}^4 \), its hash encoding at the resolution level \( l \) can be written as: \( h_{4d}(\mu, t; l) \in \mathbb{R}^m \). 
This encoded feature is a linear interpolation of the feature vectors corresponding to the vertices of the grid surrounding the insertion point. Therefore, the hash-encoded feature across \( L \) resolutions can be expressed as:
\begin{equation}
f_h(\mu,t) = [h_{4d}(\mu,t; 1),h_{4d}(\mu,t; 2),...,h_{4d}(\mu,t; L) ].
\end{equation}

We then employ a shallow, fully-fused MLP $\phi$ to cross-fuse the hash features from each resolution level.
\begin{equation}
f_d = F_d(\mu,t)  =   \phi(f_h(\mu,t)).
\end{equation}

To enable the model to adaptively balance its learning of static and dynamic residual features and accelerate convergence \cite{park2021hypernerf}, we designed a weight field \( F_w \), implemented with a shallow MLP, to predict the weights \( w_{s} \) and \( w_{d} \) for these features: \( w_{s}, w_{d} = F_w(\mu, t) \).
Given a seed and query time \( t \), we collect the outputs from the above fields and compute the weighted feature vector \( f_w \) for this seed as follows:
\begin{equation}
    f_w = w_sf_s+w_df_d
\end{equation}
where weighted feature vector $f_w$ represents the geometric information of the scene at position \( \mu \) at time \( t \).

\begin{figure}[t]
\centering
\setcounter{subfigure}{0}  
\subfloat[Static feature]{
		\includegraphics[scale=0.056]{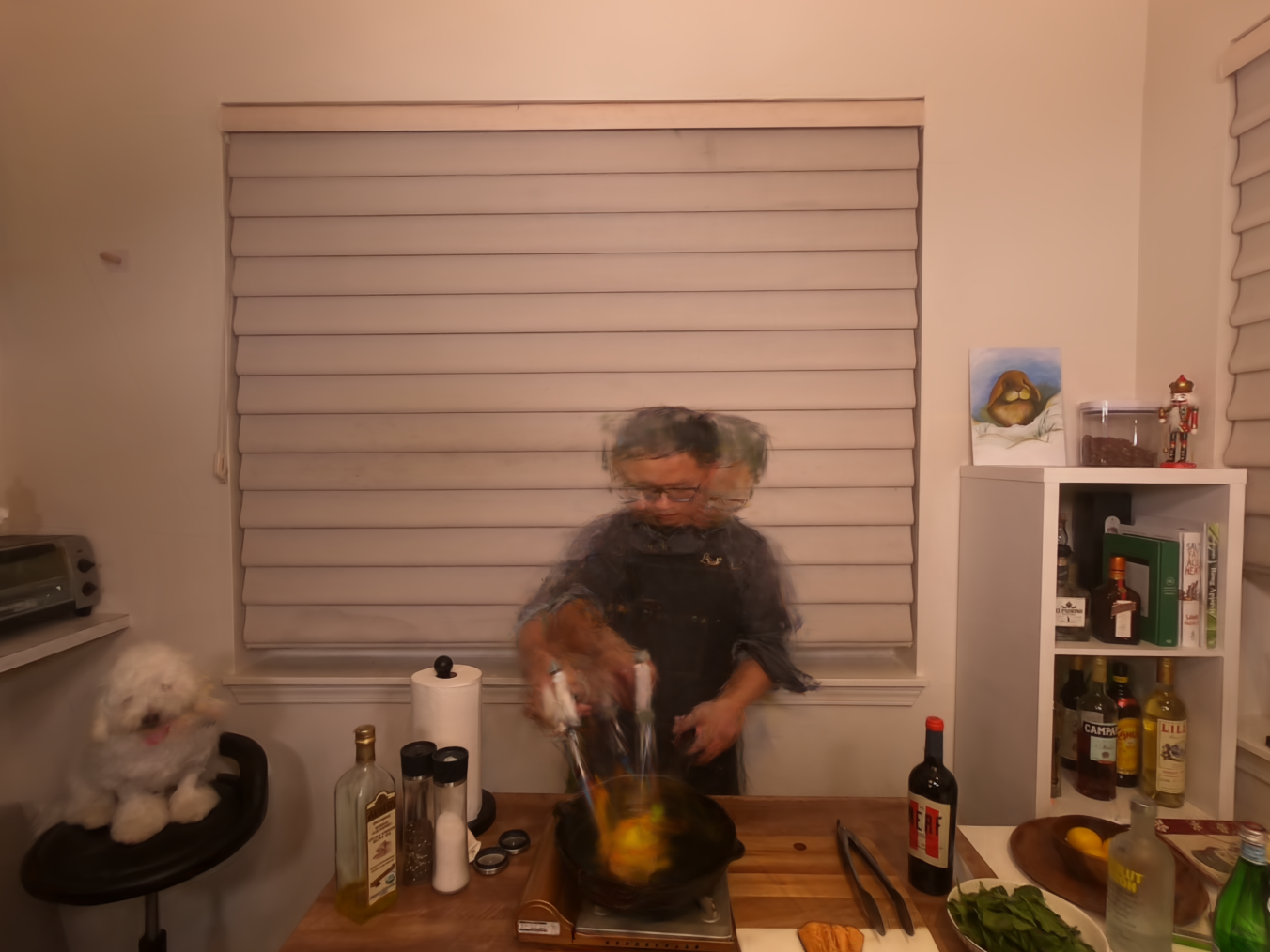}}
\subfloat[Dynamic feature]{
		\includegraphics[scale=0.056]{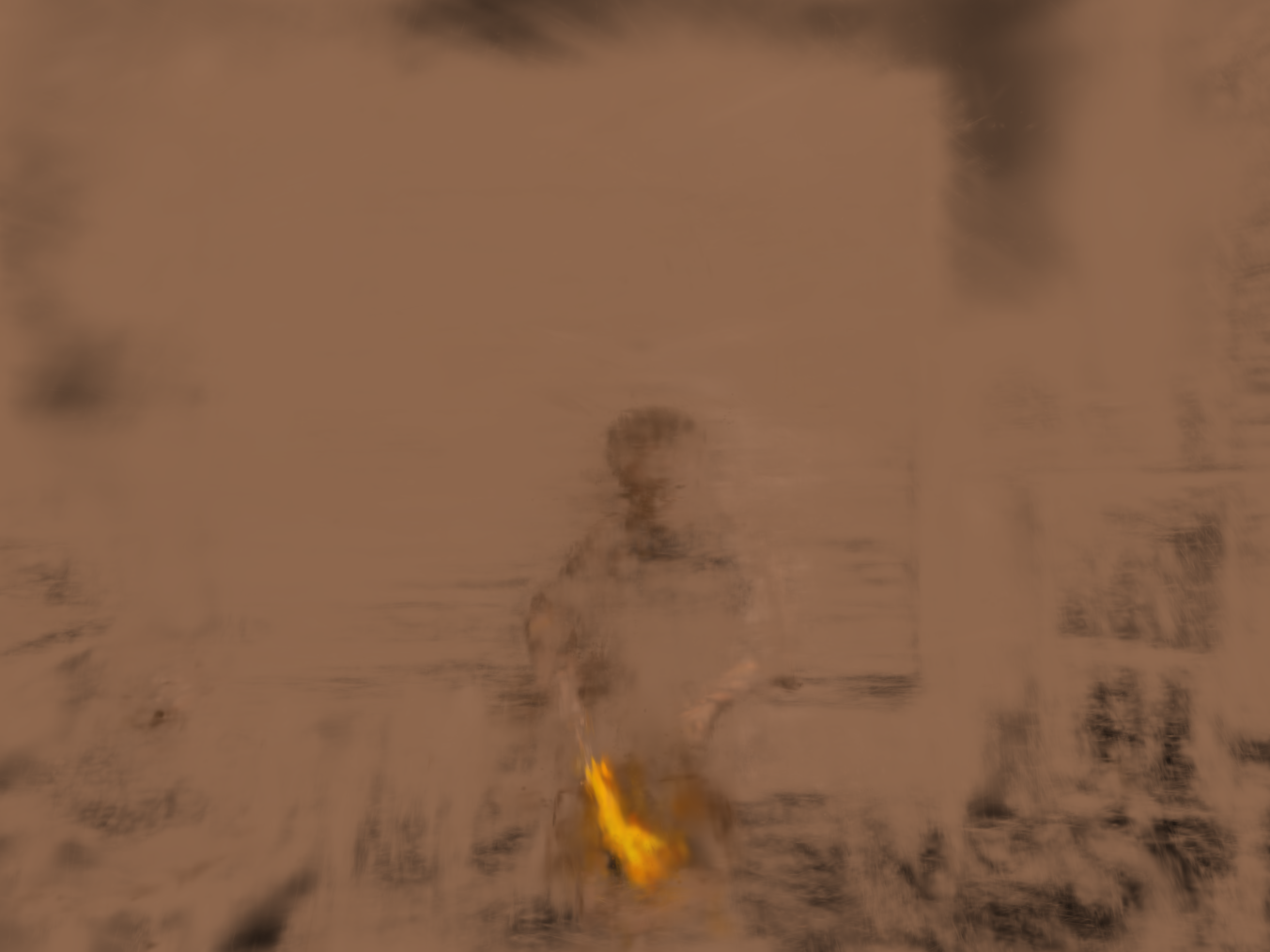}}
\subfloat[Weighted feature]{
		\includegraphics[scale=0.056]{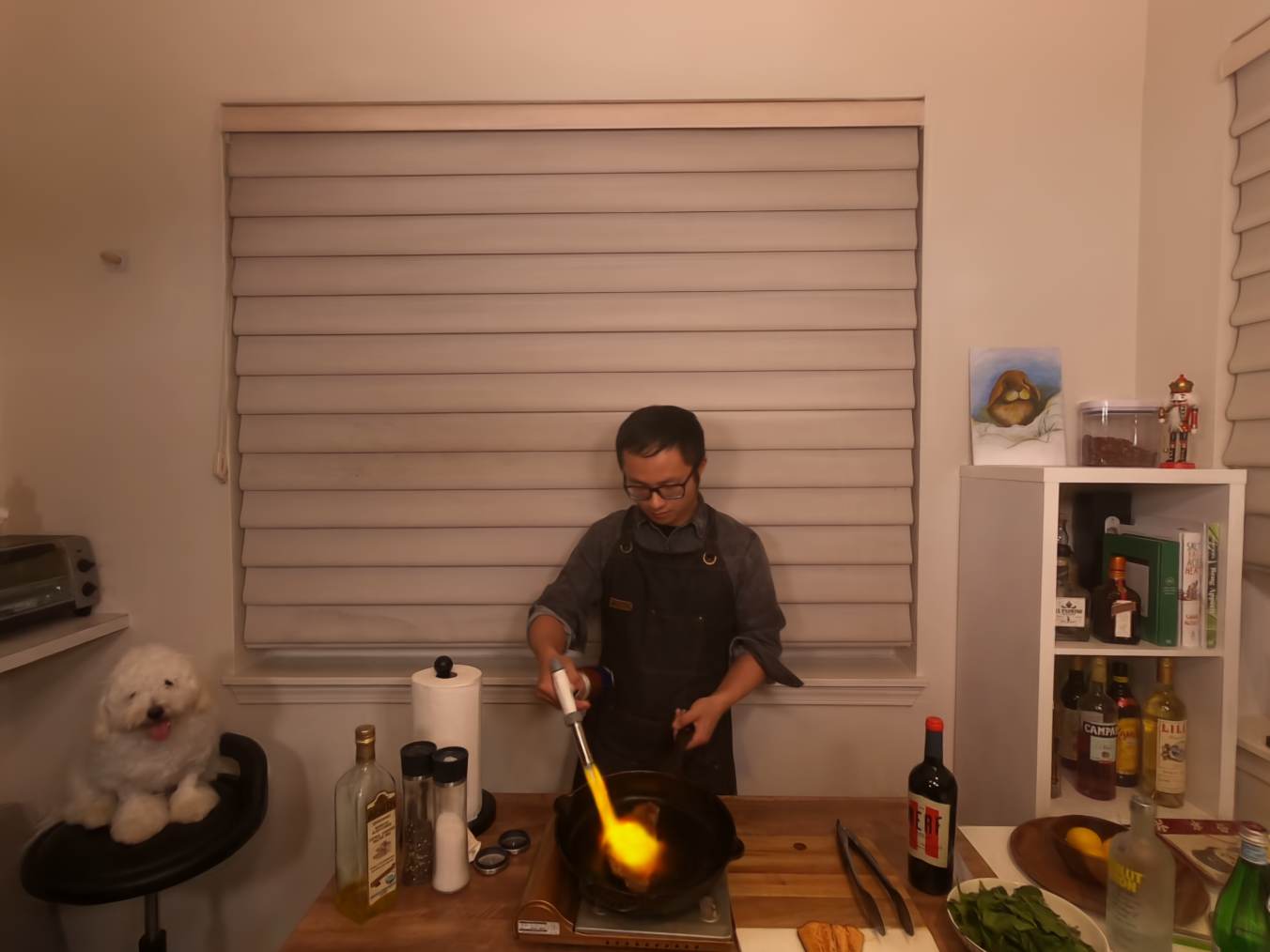}}\\
\subfloat[Distribution of dynamic residual feature value]{\includegraphics[width=0.45\textwidth, trim=0.5cm 0cm 1cm 0cm]{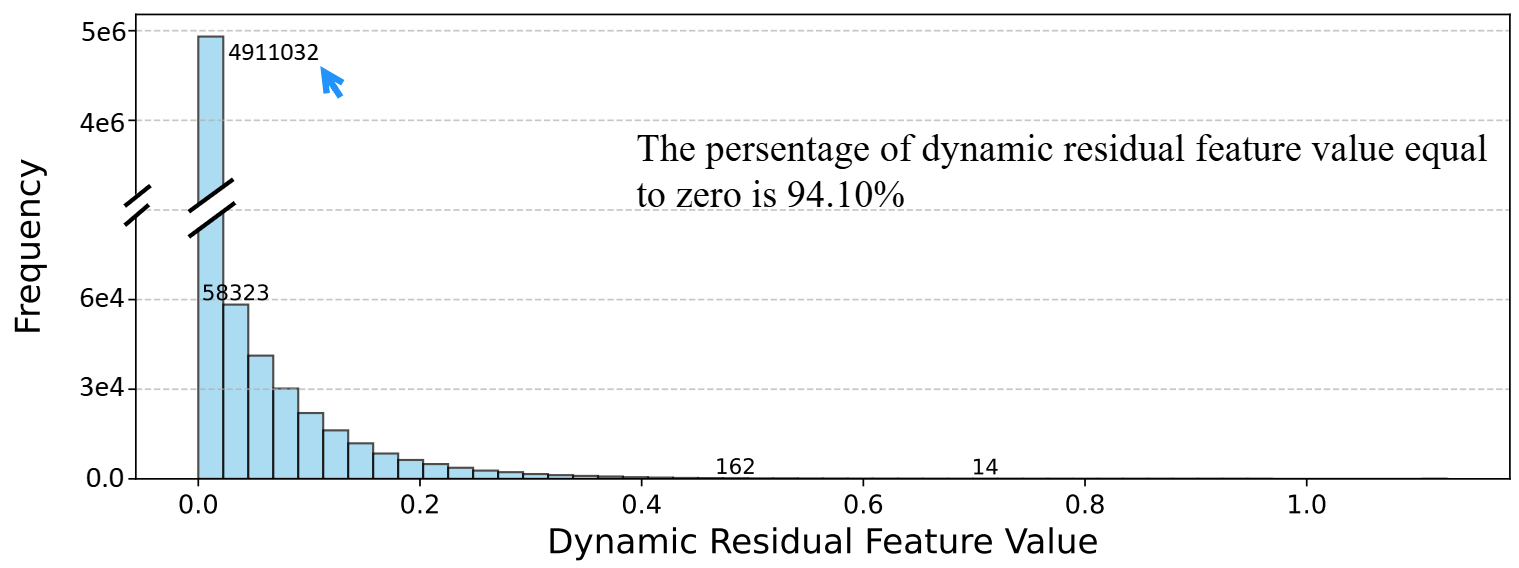}}

  \caption{(a) and (b) show the results decoded with static and dynamic residual features, demonstrating a clear separation effect; (c) shows weighted feature results; (d) shows the distribution of dynamic residual values across all local spaces. Scene information is primarily represented by static features, while dynamic residual features only capture temporal residual details; therefore, dynamic residual features tend to approach zero.} 
\label{fig: decomposition}
\vspace{-1em}
\end{figure}

In summary, the dynamic residual field supplies each local space with dynamic residual features to represent motion, which often approach zero, as shown in Fig. \ref{fig: decomposition} (d). When decoded and rendered, these features effectively capture the temporal details, as shown in Fig. \ref{fig: decomposition} (b). The predominant static information is provided by the static feature \( f_s \), depicted in Fig. \ref{fig: decomposition} (a). Together, these elements accurately represent the entire scene, as shown in Fig. \ref{fig: decomposition} (c).  This decoupling enables more effective modeling of static and dynamic components, improving rendering quality, especially in large-scale motion scenes.

\subsection{ Local Temporal Gaussian Derivation}
\label{subsec: neural gs}
In this section, we explain how to generate Temporal Gaussians from each seed as the final rendering primitives. Each Temporal Gaussian is parameterized as \( G_t\{\mu_t, q_t, s_t, \sigma_t, c_t\} \), where \( t \) denotes the query time, allowing the Temporal Gaussian to have varying parameters over time. Each seed produces \( k \) Temporal Gaussians, with their means given by:
\begin{equation}
     \{\mu_t^i\}_{i=0}^{k-1} = \mu + v \cdot F_{\mu}(f_w) 
\end{equation}
where \( \mu \) and \( v \) denote the position and scale parameters of the local space, as described in Sec. \ref{subsec: seed init}. \( \mu_t^i \) represents the \( i \)-th Temporal Gaussian generated from the seed at time \( t \), and \( F_{\mu}(\cdot) \) is a shallow MLP that outputs a vector of size \( k \times 3 \).
Inspired by \cite{lu2024scaffold}, the other Temporal Gaussian parameters are similarly predicted using individual MLPs \( F_* \). For instance, the opacity can be represented as:
\begin{equation}
   \{\sigma_t^i\}_{i=0}^{k-1} = \text{Sigmoid}( F_{o}(f_w, \mathbf{d} ) ), \quad \mathbf{d} = \frac{\mu - \mu_c}{\|\mu - \mu_c\|_2}
\end{equation}
where \( \mu_c \) is the center of the observed camera coordinates, with quaternions \( \{q_{t}^i\} \) and scales \( \{s_{t}^i\} \) similarly derived.

\textbf{Temporal Gaussian deactivation.} Through experiments, we find that some local spaces only model moving objects at the query time \( t_a \). At other times \( t_b \) (\( b \neq a \)), most Temporal Gaussians in these local spaces exhibit low opacity $\sigma_t^i$, contributing minimally to the scene representation while increasing computational load. Therefore, we set a threshold \( \tau_\alpha \) to deactivate these Temporal Gaussians, reducing computational load without affecting rendering quality. The specific ablation experiments can be found in ablation studies.

\subsection{Adaptive Seed Growing}
\label{subsec: asgs}

\begin{figure}
   \centering
   \includegraphics[width=\textwidth, trim=0cm 14cm 2cm 0.5cm]{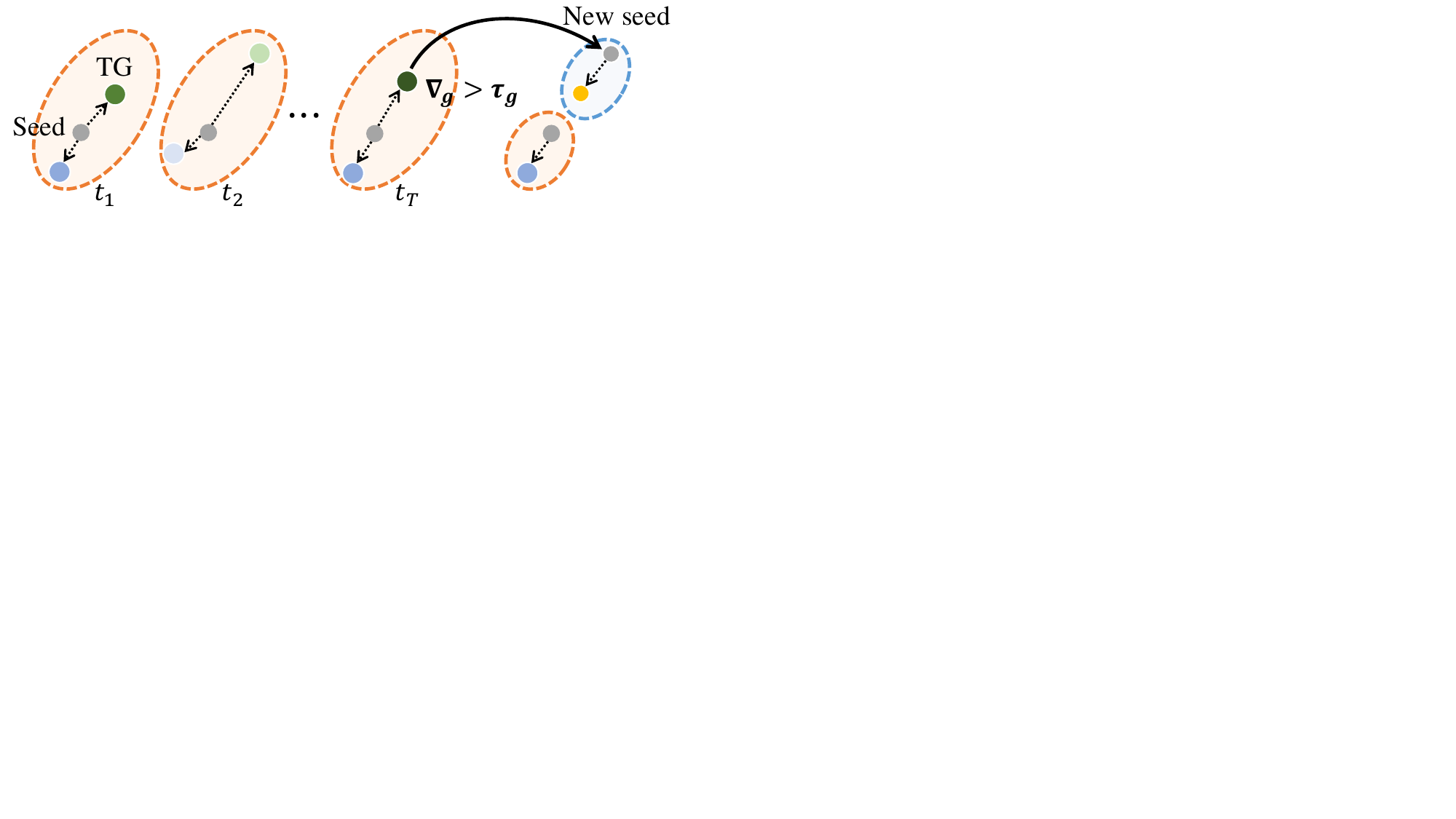}
   \caption{We add seeds where the 2D projection gradient \(\nabla_g\) of Temporal Gaussian exceeds the threshold \( \tau_{\text{g}} \) over time \(\{t_1,..,t_T\}\).}
   \label{fig: asgs}
   \vspace{-10pt}
\end{figure}

The sparse point cloud initialized by SfM often suffers from incomplete scene coverage, especially in areas with weak textures and limited observations \cite{kerbl20233d,lu2024scaffold}. This lack of coverage makes it challenging to construct precise local spaces for scene modeling, which in turn hinders convergence to high rendering quality.
To address this challenge, we propose an Adaptive Seed Growing (ASG), an error-based seed growth approach where new seeds are added in important regions identified by the Temporal Gaussians. 
As shown in Fig. \ref{fig: asgs}, within each local space, we record the maximum 2D projection gradient \( \nabla_{max}^i \) and its 3D position \( \mu_{max}^i \) for the \( i \)-th Temporal Gaussian during the \( n \) iterations. This is mathematically expressed as:
\begin{equation}
   \{ \nabla_{max}^i, \mu_{max}^i\} = \max_{t\in T}\{ \nabla_t^i, ~~~ \mu_t^i\}
\end{equation}
where \( T \) represents the set of query times corresponding to \( n \) iterations. If \( \nabla_{max}^i > \tau_g \), additional seed filling is needed, so a seed is added at \( \mu_{max}^i \) to model motion in that local space. This gradient-based growth method helps to address the limitations of the initial point cloud, enhancing the model's robustness in scene modeling. For detailed ablation studies, refer to Tab. \ref{tab: meetroom asgs ablation}.

\subsection{Loss Function}
\label{subsec:loss}

To encourage generating small Temporal Gaussians at each query time \( t \), making each responsible only for its corresponding local space, we apply a volume regularization \( L_v \), similar to that in \cite{lombardi2021mixture,lu2024scaffold}, defined as:
\begin{equation}
    L_v = \sum _{i=1}^M \text{Prod}(s_t^i)
\end{equation}
where \( M \) denotes the number of active Temporal Gaussians from all local spaces, \( \text{Prod}(\cdot) \) represents the product of the vector values, and \( s_t^i \) is the scaling of each active Temporal Gaussian at query time \( t \).
Following the 3DGS approach, we incorporate \( L_1 \) and \( L_{SSIM} \) losses to enhance reconstruction quality. The total loss function is defined as:
\begin{equation}
L = (1 - \lambda_{SSIM})L_1 + \lambda_{SSIM}L_{SSIM} + \lambda_{v}L_{v}.
\end{equation}

\section{Experiments}
\label{sec:experments}

\subsection{Implementation}
\label{subsec:Implementation}

Our method is primarily compared with current open-source SOTA methods, including SpacetimeGS~\cite{li2024spacetime}, 4DGS~\cite{wu20244d}, and 3DGStream~\cite{sun20243dgstream}. We maintain the same training iterations as 3DGS, using 30,000 iterations. For our method, we set \( k = 10 \) for all experiments, and all MLPs are 2-layer networks with ReLU activation, with the output activation function using Sigmoid or normalization. The dimensions of the dynamic and static features are set to 64. The hash table size is set to \( 2^{17} \), with other settings consistent with INGP \cite{muller2022instant}. For the ASG method, we start from 3,000 iterations to 15,000 iterations, implementing the seed point growth strategy every 100 iterations, with \( \tau_g = 0.001 \). The deactivation threshold of Temporal Gaussian is set to \( \tau_\alpha = 0.01 \). The two loss weights, \( \lambda_{SSIM} \) and \( \lambda_{vol} \), are set to 0.2 and 0.001, respectively, with the optimizer being Adam \cite{kingma2014adam}, following the learning rate of 3DGS~\cite{kerbl20233d}. All experiments are conducted on an NVIDIA RTX 3090 GPU.

\subsection{Datasets}
\label{subsec:datasets}

We primarily evaluate our method on the fine-scale motion datasets N3DV \cite{li2022neural} and MeetRoom \cite{li2022streaming}, consistent with most multi-view methods \cite{sun20243dgstream, wang2023mixed, li2022streaming}. To further evaluate the robustness of our method in large-scale dynamic scenes, we test our method on more challenging VRU basketball court dataset \cite{wuswift4d}. 

\textbf{The N3DV dataset \cite{li2022neural}} is a widely used benchmark, captured by a multi-view system of 21 cameras, recording dynamic scenes at a resolution of \( 2704 \times 2028 \) and 30 FPS. Following previous work~\cite{li2022neural,wu20244d,sun20243dgstream}, we downsample the dataset and split cameras for training and testing.

\textbf{The MeetRoom dataset \cite{li2022streaming}} is even more challenging, captured by a multi-view system with only 13 cameras, recording dynamic scenes at a resolution of \( 1280 \times 720 \) and 30 FPS. In line with prior work \cite{sun20243dgstream,li2022streaming}, we use 12 cameras for training and reserve one for testing.

\textbf{The VRU Basketball Court dataset \cite{VRU}} is captured using a 34-camera multi-view system, recording real-world basketball games \textit{GZ}, \textit{DG4} at \( 1920 \times 1080 \) resolution and 25 FPS.  We use 30 cameras for training, reserving 4 cameras (cameras 0, 10, 20, 30) for testing. This dataset is used for the first time in Swift4D \cite{wuswift4d}, and is provided by AVS-VRU \cite{AVS}  for academic use. Compared to previous \textbf{fine-scale} motion datasets \cite{li2022neural, li2022streaming}, it features \textbf{larger motion scales} and better evaluates the dynamic modeling capability of the dynamic methods.

\begin{table}
\centering

\caption{
    \textbf{Quantitative comparisons on the Neural 3D Video Dataset \cite{li2021neural}.}
    ``Size'' is the total model size for 300 frames. DSSIM$_1$ sets data range to 1.0 while DSSIM$_2$ to 2.0 \cite{li2024spacetime}. $\ast$ indicates online method.
}
\label{tab: n4d}
\resizebox{\linewidth}{!}{
\begin{tabular}{@{}lccccccc}
    \toprule
   Method & PSNR$\uparrow$ & DSSIM$_1$$\downarrow$& DSSIM$_2$$\downarrow$ & LPIPS$\downarrow$ & FPS$\uparrow$ & Time$\downarrow$ & Size$\downarrow$ \\
  \midrule
  StreamRF~$^\ast$~\cite{li2022streaming}  & 28.26 & - & - & - & 10.9 & - & 5310 MB \\
  NeRFPlayer~\cite{song2023nerfplayer}  & 30.69 & 0.034 & - & 0.111 & 0.05 & 6.0 h & 5130 MB \\
  HyperReel~\cite{attal2023hyperreel}  & 31.10 & 0.036 & - & 0.096 & 2 & - & 360 MB \\
  K-Planes~\cite{fridovich2023k}    &  31.63 & - &\cellcolor{yellow!25}0.018  & - & 0.3  & 5.0 h & 311 MB \\
  HexPlane~\cite{cao2023hexplane}    &  31.70 & - &\cellcolor{red!25}0.014  & 0.075 & 0.21  & 12.0 h & 240 MB \\
  MixVoxels~\cite{wang2022mixed}   & 31.73 &  - &\cellcolor{orange!25}0.015   & 0.064 & 4.6 & - & 500 MB \\
  4DGaussian~\cite{wu20244d}     & 31.02 & \cellcolor{yellow!25}0.030  & - & 0.150  & 30  & \cellcolor{orange!25}0.67 h & \cellcolor{red!25}90 MB \\
  3DGStream$^1$ ~\cite{sun20243dgstream}     & 31.67 & -  & - & -  & \cellcolor{red!25}215  & \cellcolor{yellow!25}1.0 h & 1230 MB \\
 RealTimeGS~\cite{yang2023real}     &  \cellcolor{yellow!25}32.01 & -  & \cellcolor{red!25}0.014 & \cellcolor{yellow!25}0.055  & \cellcolor{yellow!25}114  & 9.0 h &  $>1000$ MB  \\
  SpaceTimeGS ~\cite{li2024spacetime}         & \cellcolor{orange!25}32.05 & \cellcolor{red!25}0.026  & \cellcolor{red!25}0.014 & \cellcolor{orange!25}0.044  & \cellcolor{orange!25}140 & $>5$ h & \cellcolor{yellow!25}200 MB \\
   \textbf{\ourname(Ours)}         & \cellcolor{red!25}32.28 & \cellcolor{orange!25}{0.028}  & \cellcolor{red!25}{0.014} & \cellcolor{red!25}{0.043}  &105 & \cellcolor{red!25}{0.58 h} & \cellcolor{orange!25}{100 MB} \\
  \bottomrule
\end{tabular}
}
\vspace{-1em}
\end{table}

\begin{table}[t]
    \centering
        \caption{\textbf{Quantitative comparison on the MeetRoom dataset \cite{li2022streaming}.} PSNR is averaged across all 300 frames, while training time and storage requirements accumulate over the entire sequence. }
    \begin{threeparttable}
    \resizebox{0.8\columnwidth}{!}
    {
        \begin{tabular}{lcccr@{}}
        \toprule
         Method & PSNR \(\uparrow\)  & Time(hours) \(\downarrow\)  & Size(MB) \(\downarrow\) \\
        \midrule[0.5pt]
        \multirow{1}{*}{Plenoxel \cite{fridovich2022plenoxels}}  & 27.15 & 70 & 304500 \\
        \multirow{1}{*}{I-NGP \cite{muller2022instant}}  & 28.10 & 5.5 & 14460 \\
        \multirow{1}{*}{3DGS \cite{kerbl20233d}}  & 31.31 & 13 & 6330 \\
        \midrule[0.5pt]
        \multirow{1}{*}{StreamRF \cite{li2022streaming}}  & 26.72 & 0.85 & 2700 \\
        \multirow{1}{*}{3DGStream \cite{sun20243dgstream}}  & 30.79& 0.6 & 1230 \\
        \rowcolor{gray!20} 
        \multirow{1}{*}{\textbf{\ourname(Ours)}}  & \textbf{32.45} & \textbf{0.36} & \textbf{90}  \\
        \bottomrule
        \end{tabular}
    }
\end{threeparttable}
    \label{tab: meetroom all quality comparison}
\end{table}

\begin{table}[t]
    \centering
    \caption{\textbf{Quantitative comparison on VRU (GZ) basketball court dataset \cite{VRU}.} Static methods are tested on frame 0.}
    \begin{threeparttable}
    \resizebox{0.8\linewidth}{!}
    {
        \begin{tabular}{lcccccccccccc}
        \toprule
         Method & PSNR \(\uparrow\)  & SSIM \(\uparrow\)  & LPIPS \(\downarrow\) \\
        \midrule[0.5pt]
        \multirow{1}{*}{GOF \cite{yu2024gaussian} }  & 30.39 & 0.949 & 0.141 \\
        \multirow{1}{*}{2DGS \cite{huang20242d} }  & 30.78 & 0.949 & 0.187 \\
        \multirow{1}{*}{3DGS \cite{kerbl20233d}}  & 30.50 & 0.949 & 0.171 \\
        \midrule[0.5pt]
        \multirow{1}{*}{4DGS \cite{wu20244d} }  & 28.32 & 0.930 & 0.186 \\
        \multirow{1}{*}{SpaceTimeGS \cite{li2024spacetime} }  &   27.42 &   0.926  &  0.193 \\
        \rowcolor{gray!20} 
        \multirow{1}{*}{\textbf{\ourname}(Ours)}  & \textbf{30.58} & \textbf{0.944} & \textbf{0.173}  \\
        \bottomrule
        \end{tabular}
}

\end{threeparttable}
    \label{tab: vru compare}
\vspace{-1em}
\end{table}

\begin{figure*}[t]
    \centering
    
    \begin{minipage}[b]{0.25\textwidth}
        \centering
        \includegraphics[width=0.99\linewidth]{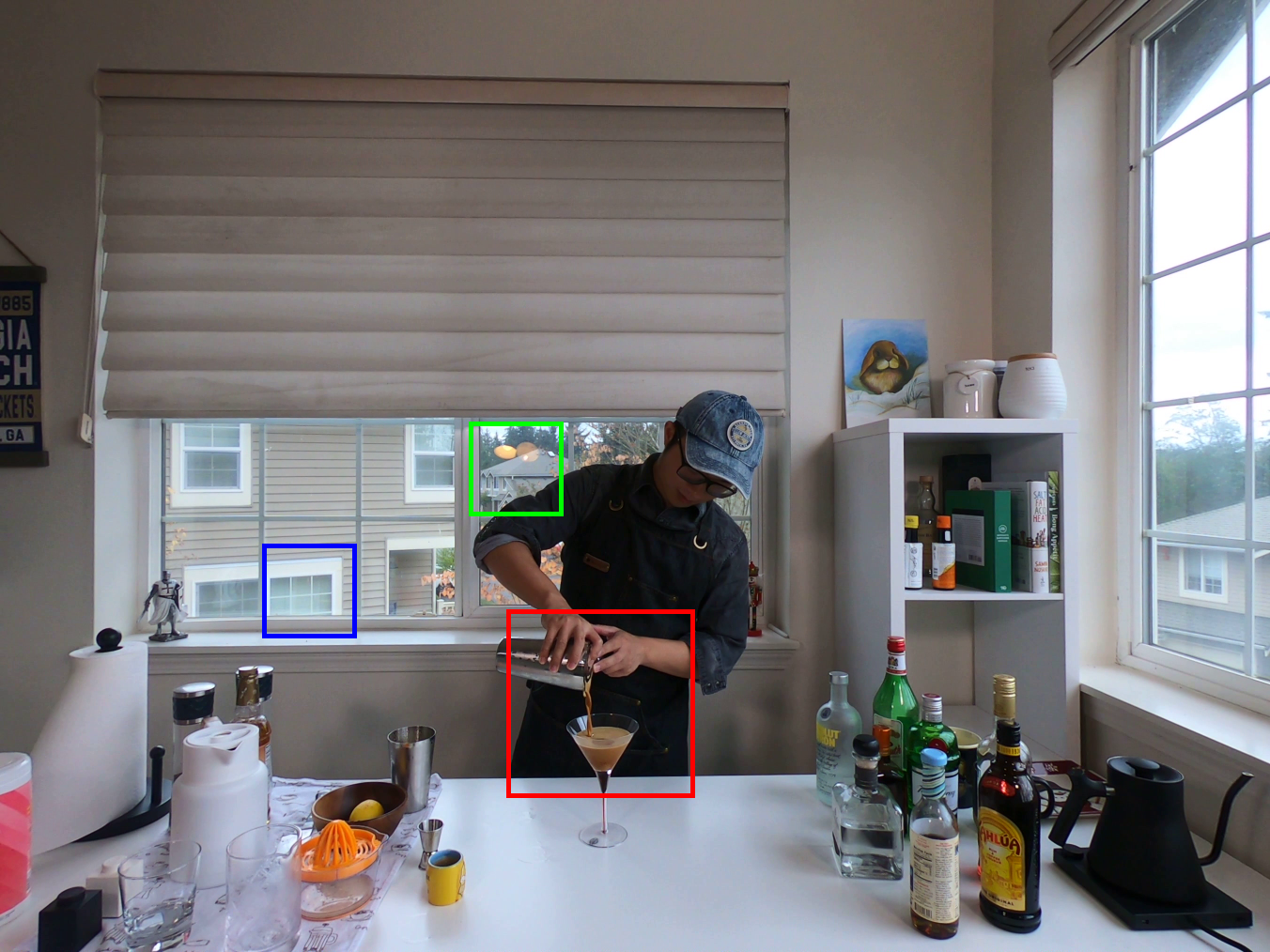} \\
        \includegraphics[width=0.32\linewidth]{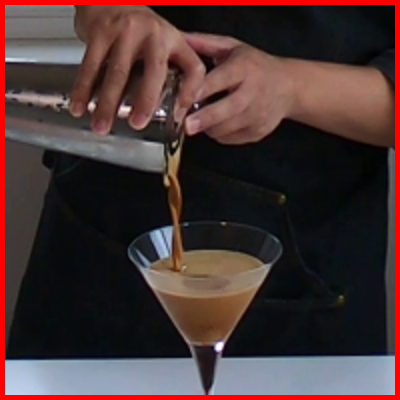}
         \hspace{-0.03\linewidth} 
        \includegraphics[width=0.32\linewidth]{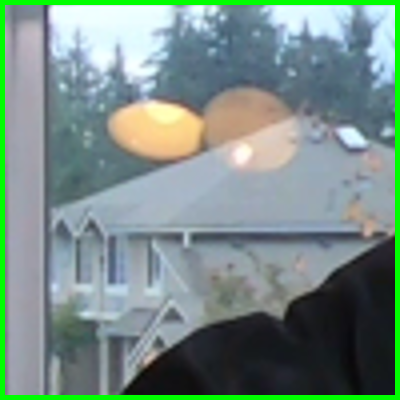}
         \hspace{-0.03\linewidth} 
        \includegraphics[width=0.32\linewidth]{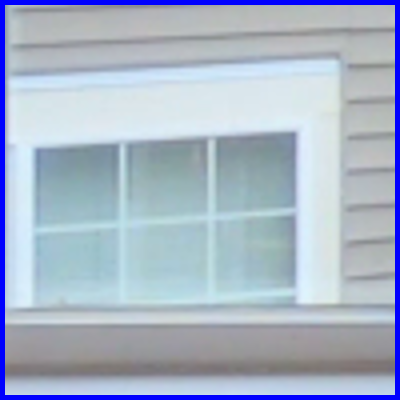}
    \end{minipage}%
    \hfill
    \begin{minipage}[b]{0.25\textwidth}
        \centering
        \includegraphics[width=0.99\linewidth]{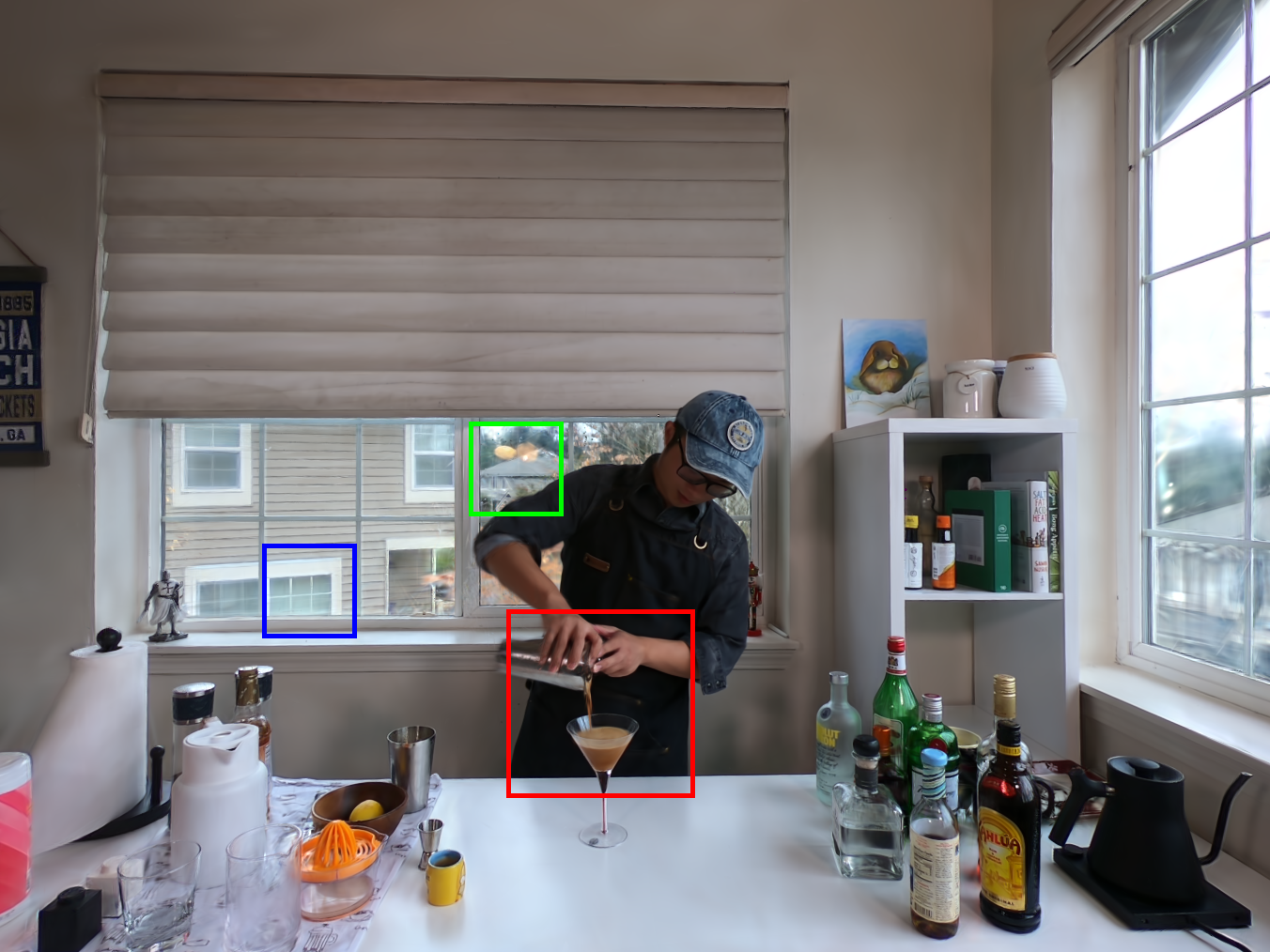} \\
        \includegraphics[width=0.32\linewidth]{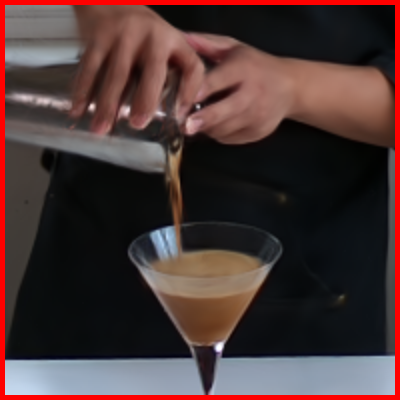}
        \hspace{-0.03\linewidth} 
        \includegraphics[width=0.32\linewidth]{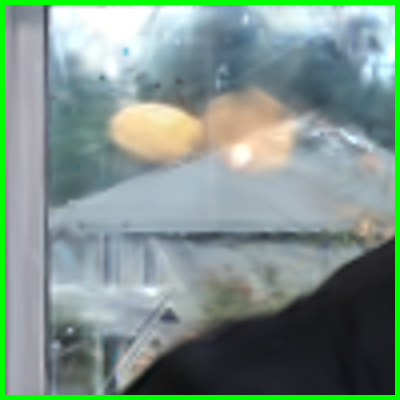}
        \hspace{-0.03\linewidth} 
        \includegraphics[width=0.32\linewidth]{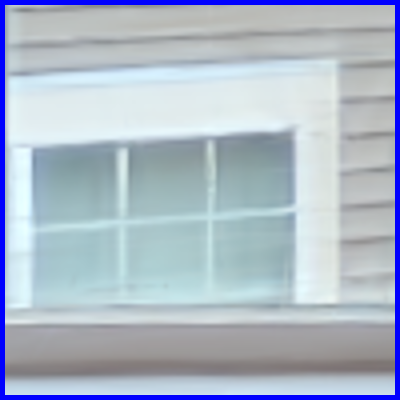}
    \end{minipage}%
    \hfill
    \begin{minipage}[b]{0.25\textwidth}
        \centering
        \includegraphics[width=0.99\linewidth]{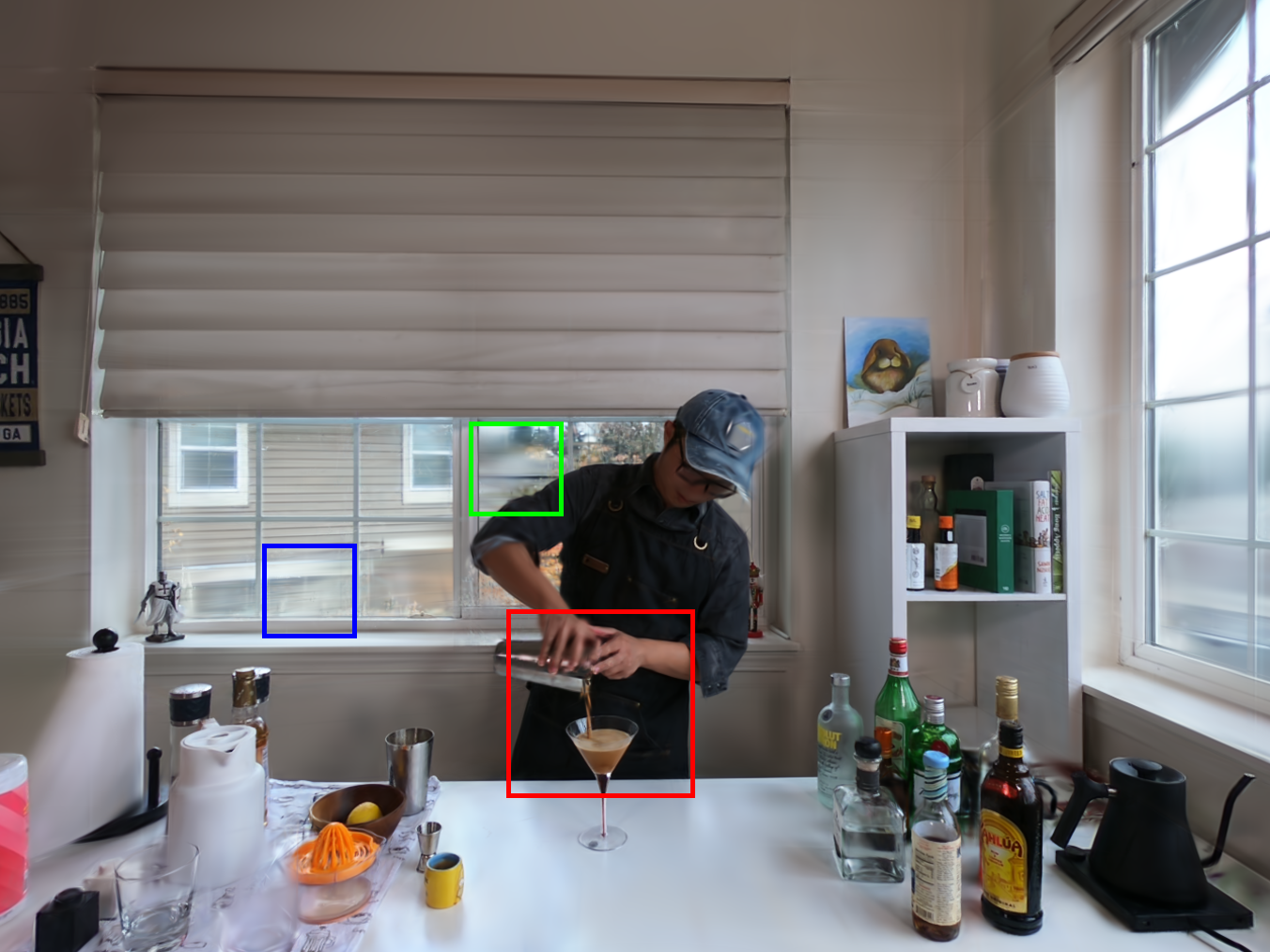} \\
        \includegraphics[width=0.32\linewidth]{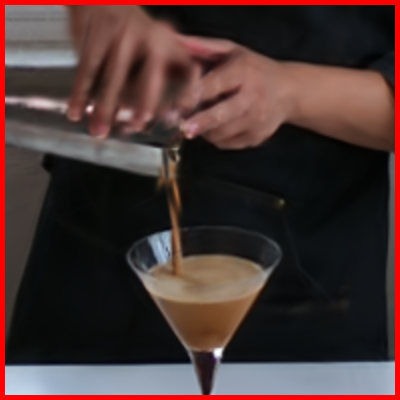}
        \hspace{-0.03\linewidth} 
        \includegraphics[width=0.32\linewidth]{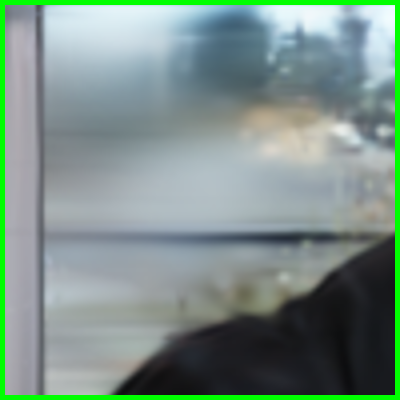}
        \hspace{-0.03\linewidth} 
        \includegraphics[width=0.32\linewidth]{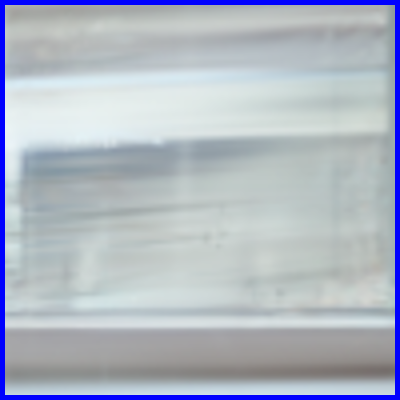}
    \end{minipage}%
    \hfill
    \begin{minipage}[b]{0.25\textwidth}
        \centering
        \includegraphics[width=0.99\linewidth]{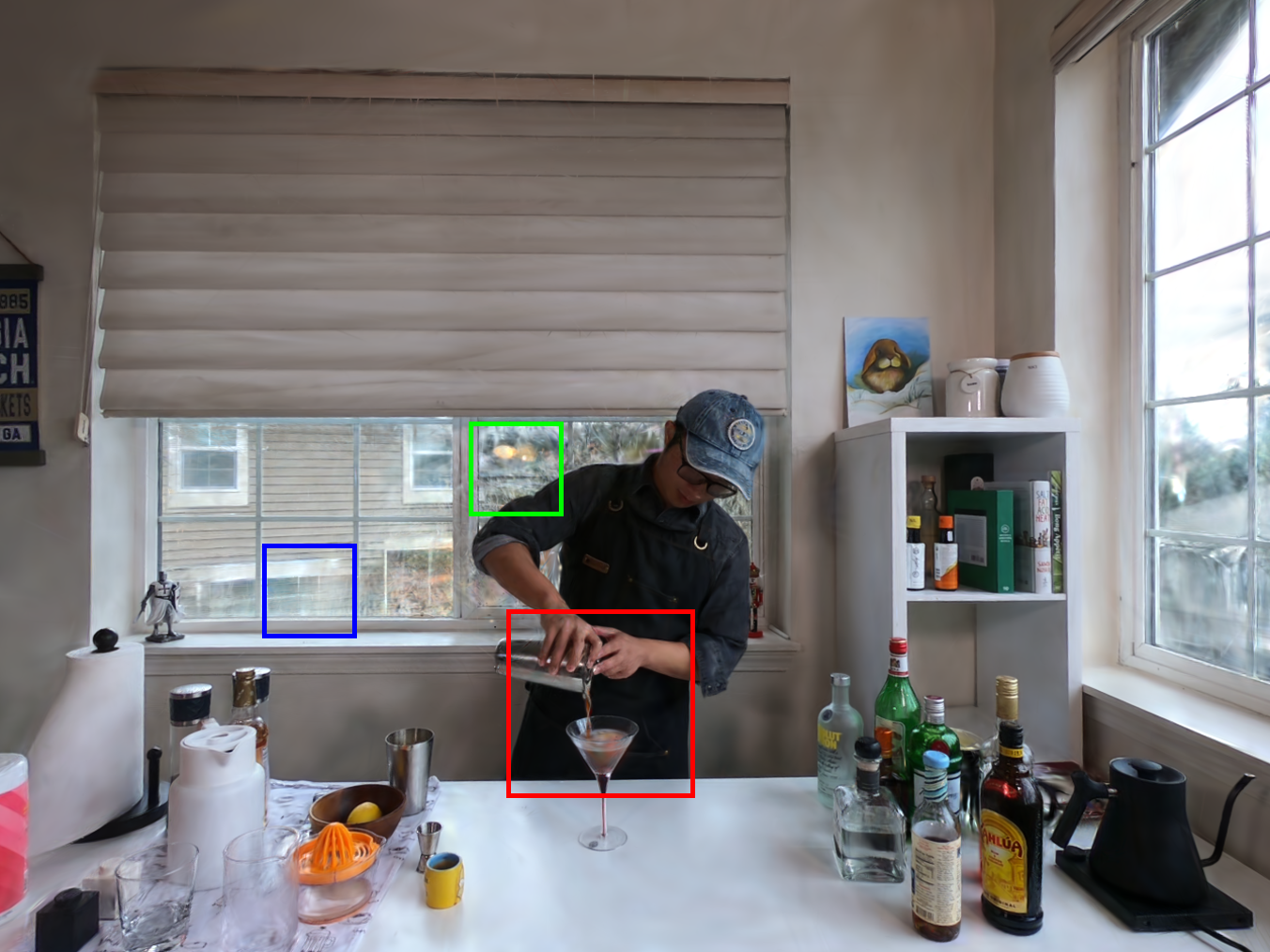} \\
        \includegraphics[width=0.32\linewidth]
        {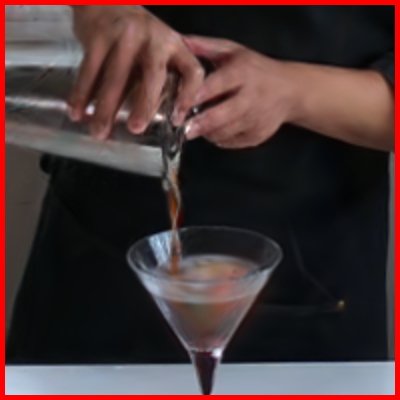}
        \hspace{-0.03\linewidth} 
        \includegraphics[width=0.32\linewidth]{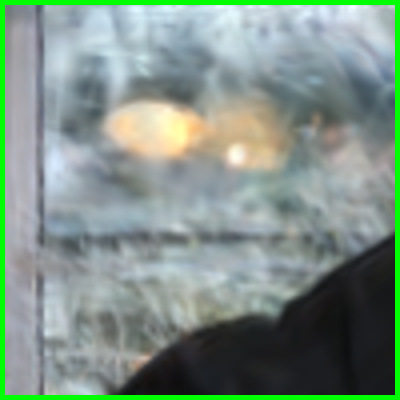}
        \hspace{-0.03\linewidth} 
        \includegraphics[width=0.32\linewidth]{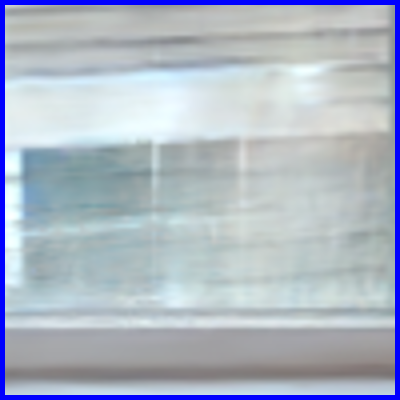}
    \end{minipage}

    \begin{minipage}[b]{0.25\textwidth}
        \centering
        \includegraphics[width=0.99\linewidth]{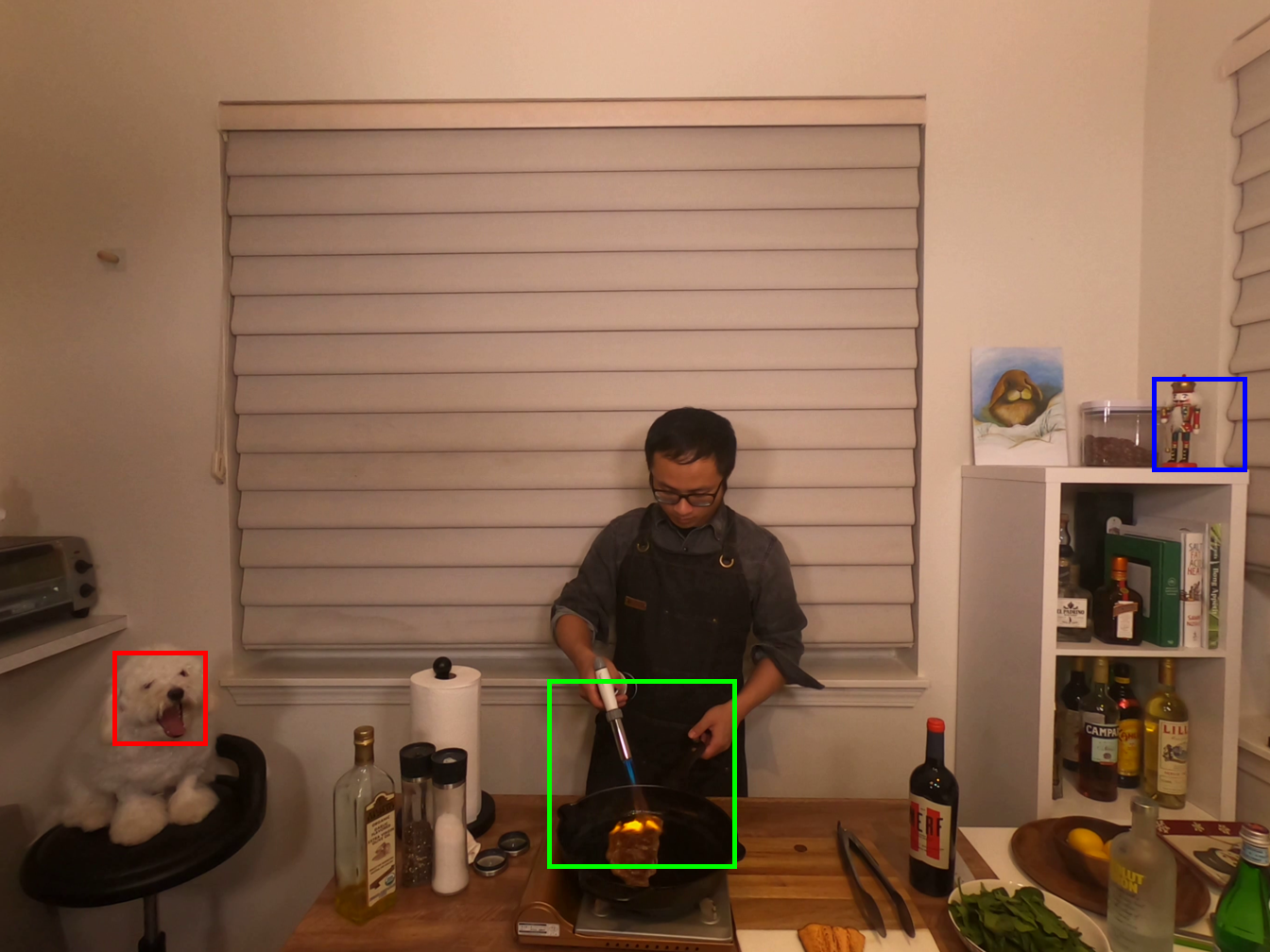} \\
        \includegraphics[width=0.32\linewidth]{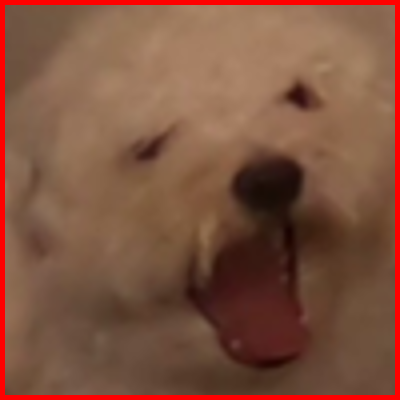}
        \hspace{-0.03\linewidth} 
        \includegraphics[width=0.32\linewidth]{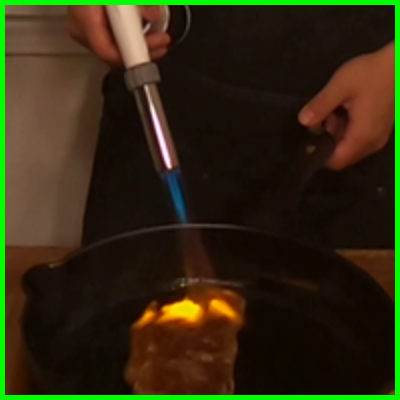}
        \hspace{-0.03\linewidth} 
        \includegraphics[width=0.32\linewidth]{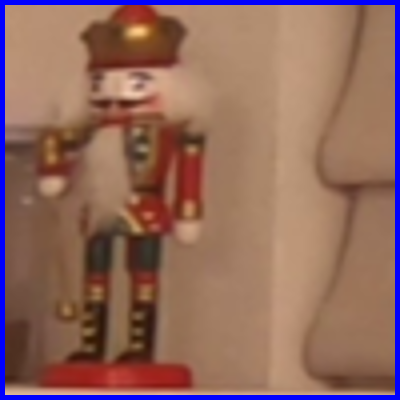}
        \caption*{(a) GT}
    \end{minipage}%
    \hfill
    \begin{minipage}[b]{0.25\textwidth}
        \centering
        \includegraphics[width=0.99\linewidth]{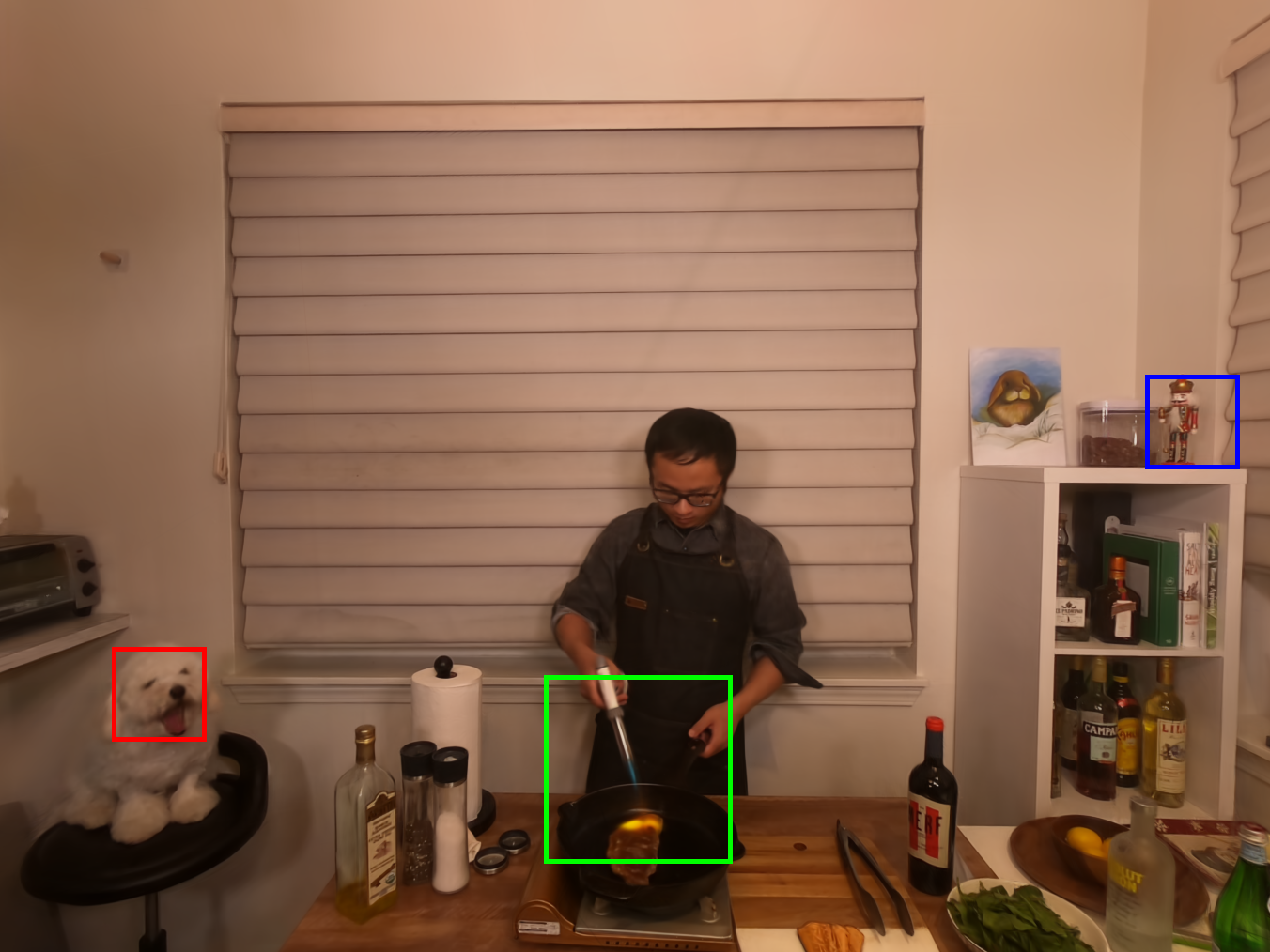} \\
        \includegraphics[width=0.32\linewidth]{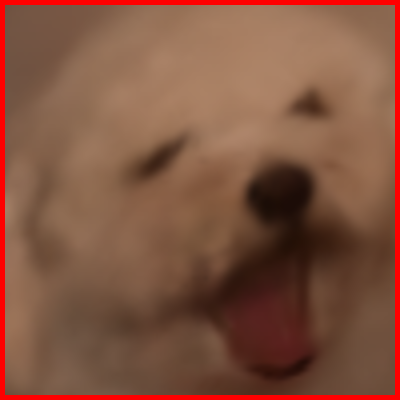}
        \hspace{-0.03\linewidth} 
        \includegraphics[width=0.32\linewidth]{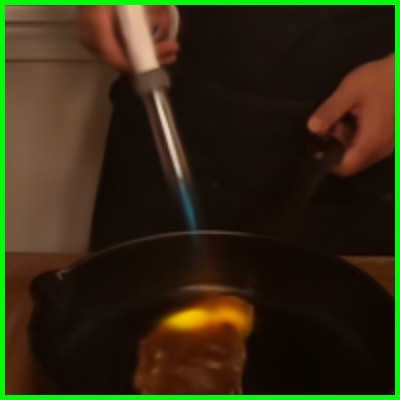}
        \hspace{-0.03\linewidth} 
        \includegraphics[width=0.32\linewidth]{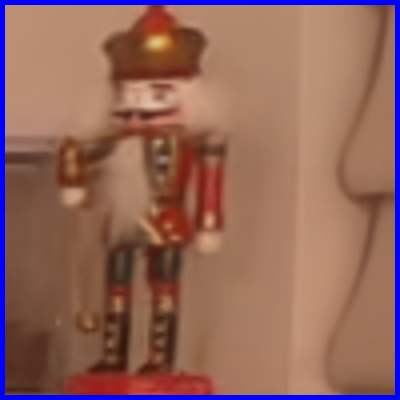}
        \caption*{(b) Ours}
    \end{minipage}%
    \hfill
    \begin{minipage}[b]{0.25\textwidth}
        \centering
        \includegraphics[width=0.99\linewidth]{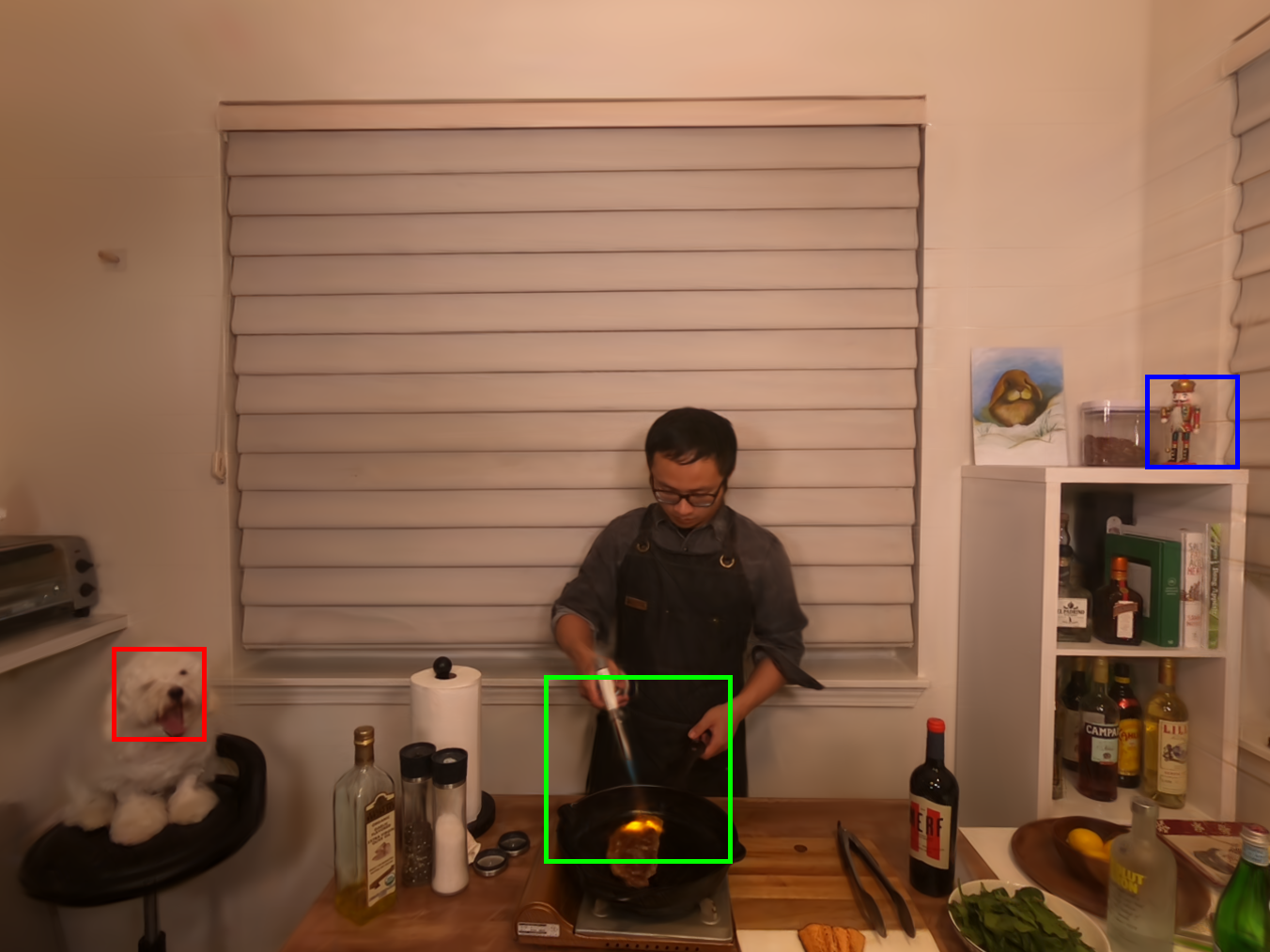} \\
        \includegraphics[width=0.32\linewidth]{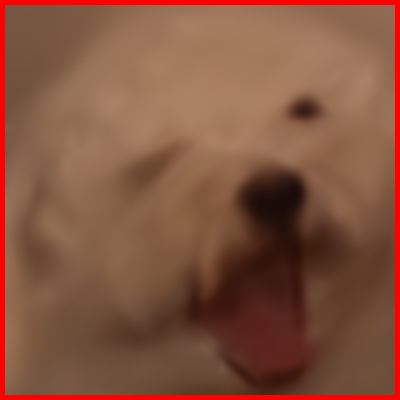}
        \hspace{-0.03\linewidth} 
        \includegraphics[width=0.32\linewidth]{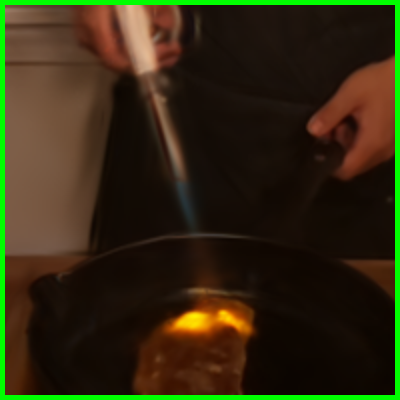}
        \hspace{-0.03\linewidth} 
        \includegraphics[width=0.32\linewidth]{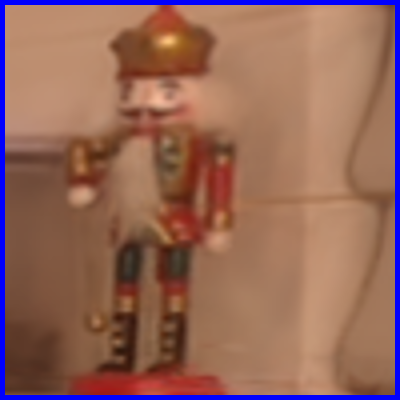}
        \caption*{(c) SpaceTimeGS \cite{li2024spacetime}}
    \end{minipage}%
    \hfill
    \begin{minipage}[b]{0.25\textwidth}
        \centering
        \includegraphics[width=0.99\linewidth]{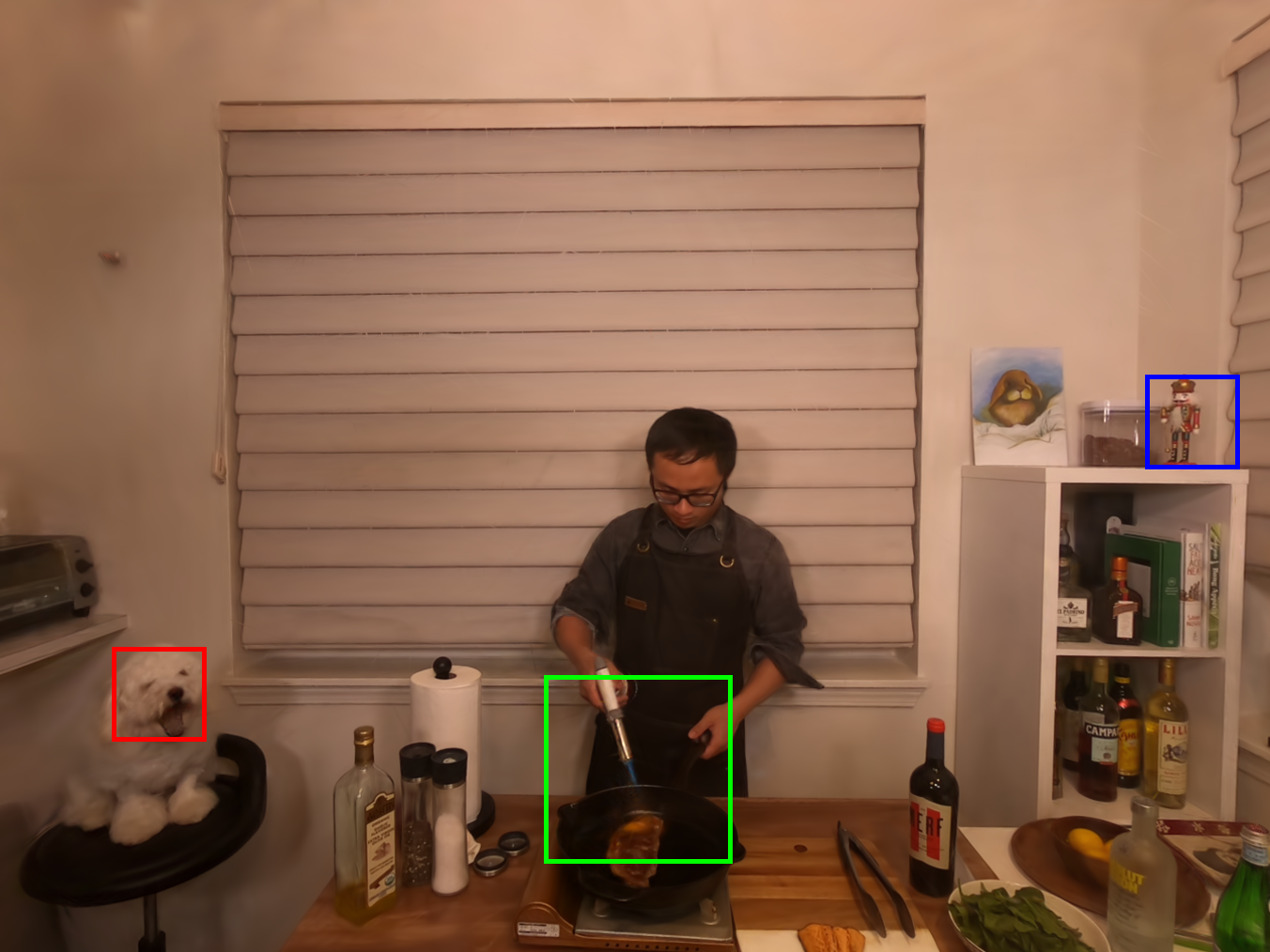} \\
        \includegraphics[width=0.32\linewidth]{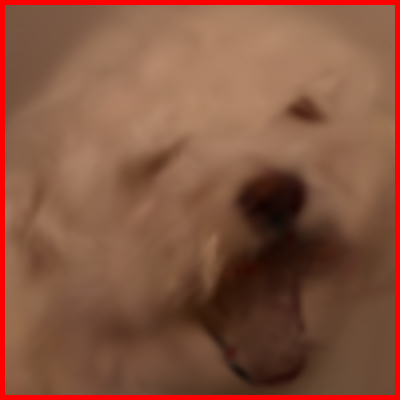}
        \hspace{-0.03\linewidth} 
        \includegraphics[width=0.32\linewidth]{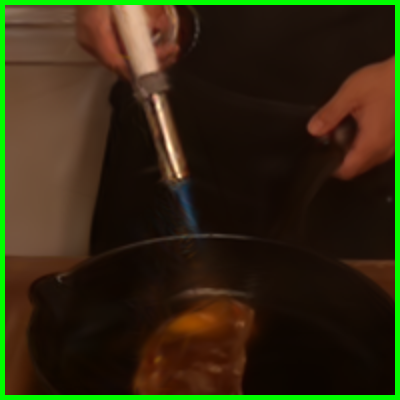}
        \hspace{-0.03\linewidth} 
        \includegraphics[width=0.32\linewidth]{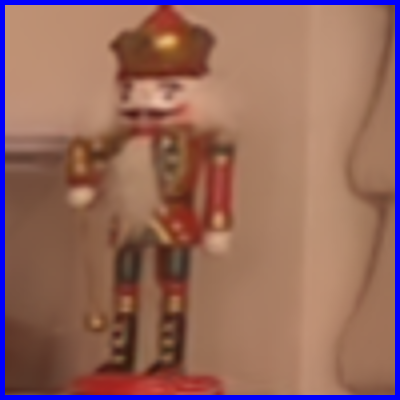}
        \caption*{(d) 3DGStream \cite{sun20243dgstream}}
    \end{minipage}

    \caption{ Qualitative results of \textit{coffee martini} and \textit{sear steak} from the N3DV dataset \cite{li2022neural} (a dataset featuring fine-scale motion). We compare our method with SOTA approaches, including STGS \cite{li2024spacetime} and 3DGStream \cite{sun20243dgstream}. Our method produces fewer floaters and preserves more details in the dynamic scene, such as newly appearing objects (e.g., coffee liquid and flame), distant background elements, and the dog’s face. }
    \label{fig:nv3d_comparison}
\end{figure*}
\begin{figure*}[t]
\centering
\setcounter{subfigure}{0}  
\subfloat[GT]{
		\includegraphics[scale=0.1]{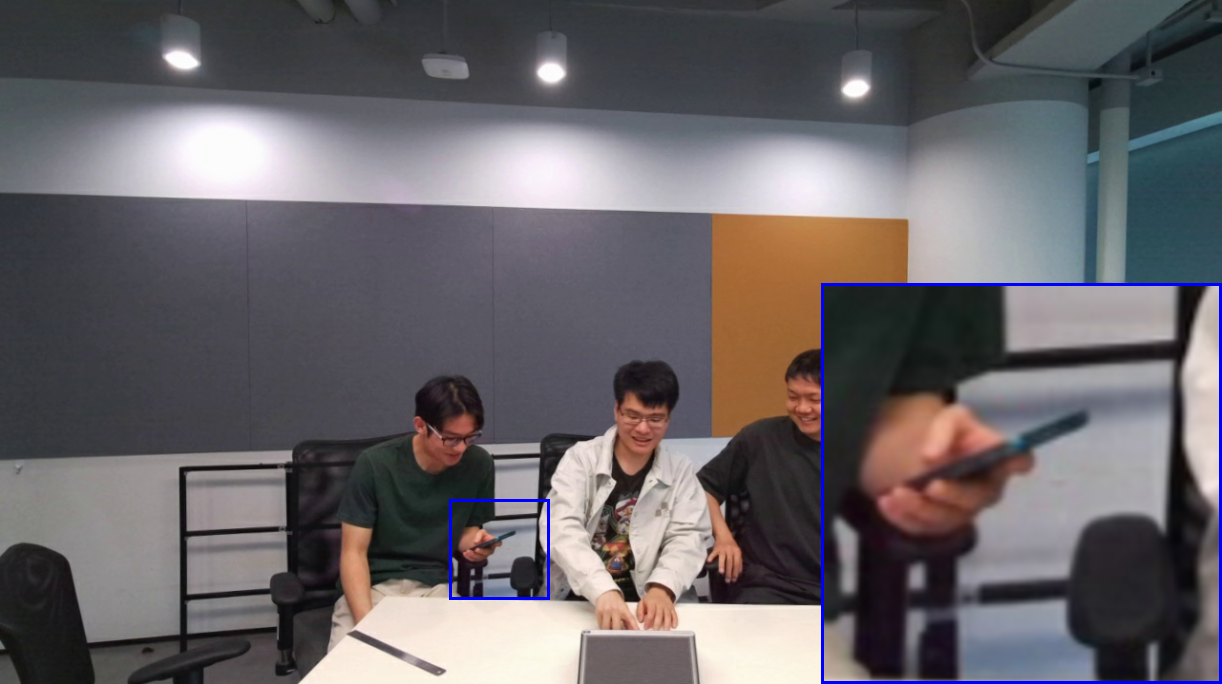}}
  \subfloat[Ours]{
		\includegraphics[scale=0.1]{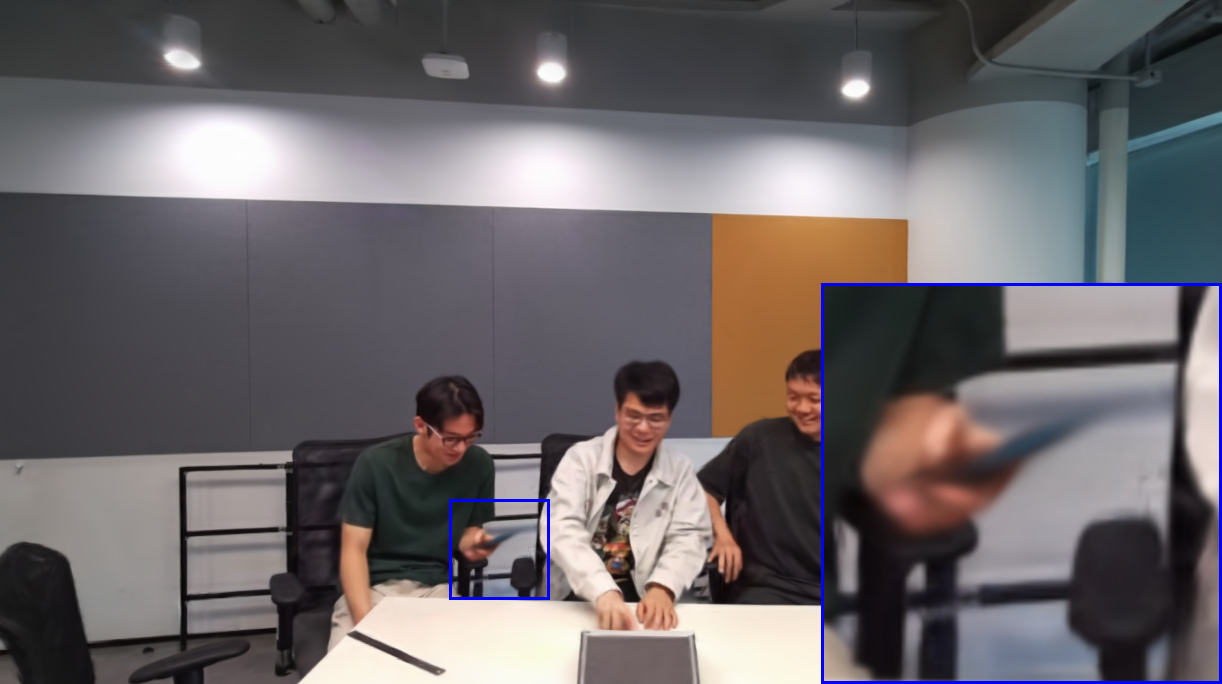}}
\subfloat[3DGStream \cite{sun20243dgstream}]{
		\includegraphics[scale=0.1]{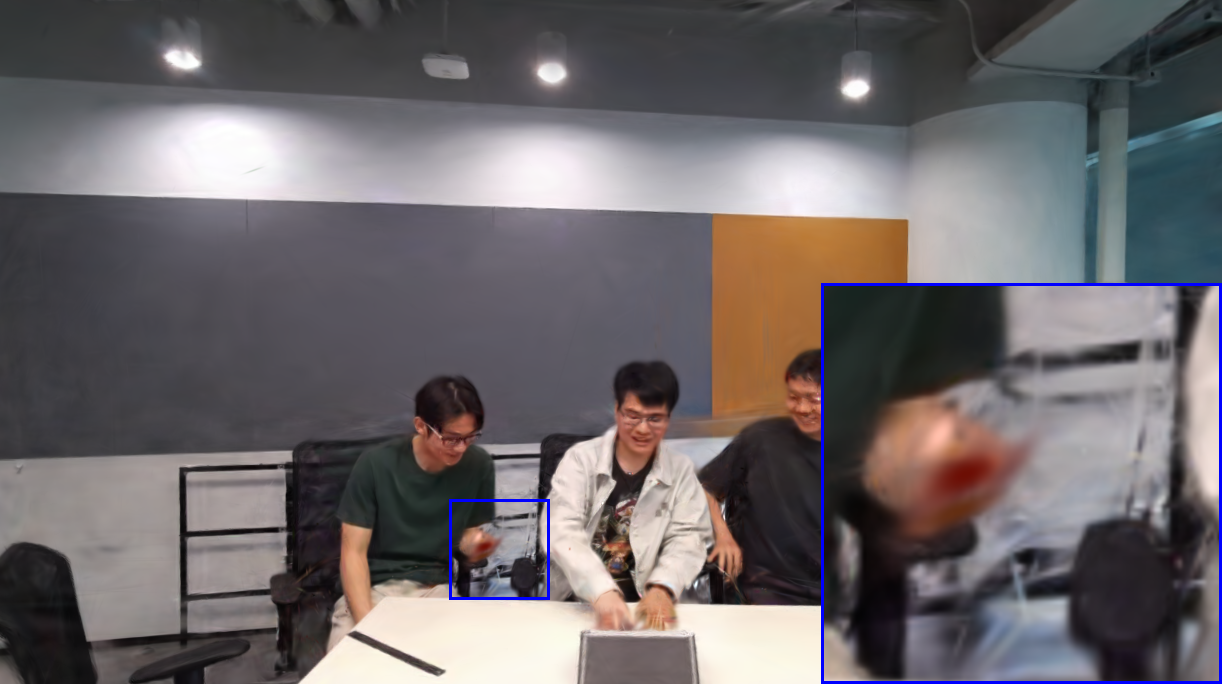}}
\subfloat[3DGS \cite{kerbl20233d}]{
		\includegraphics[scale=0.1]{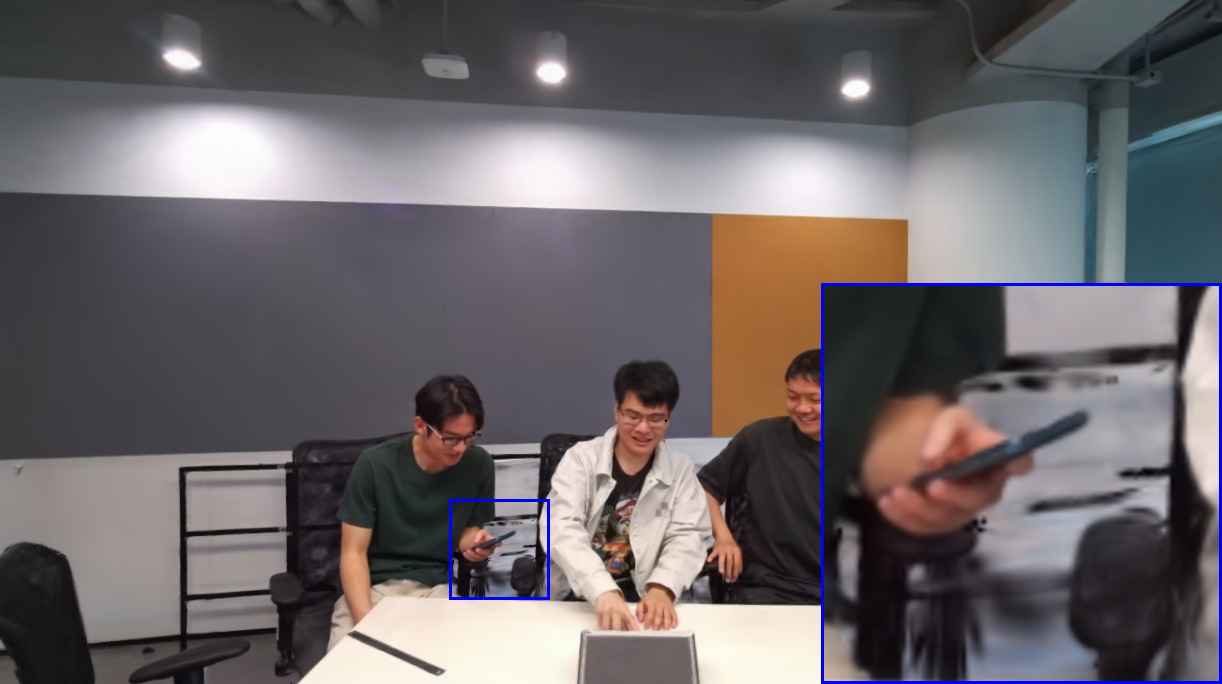}}
\caption{ Qualitative result on the \textit{discussion} of Meetroom dataset \cite{li2022streaming} (a dataset featuring sparse views and large textureless regions).}
\label{fig: meetroom compairl}
\end{figure*}

\begin{figure*}[t]
    \centering

    \begin{minipage}[b]{0.25\textwidth}
        \centering
        \includegraphics[width=0.98\linewidth]{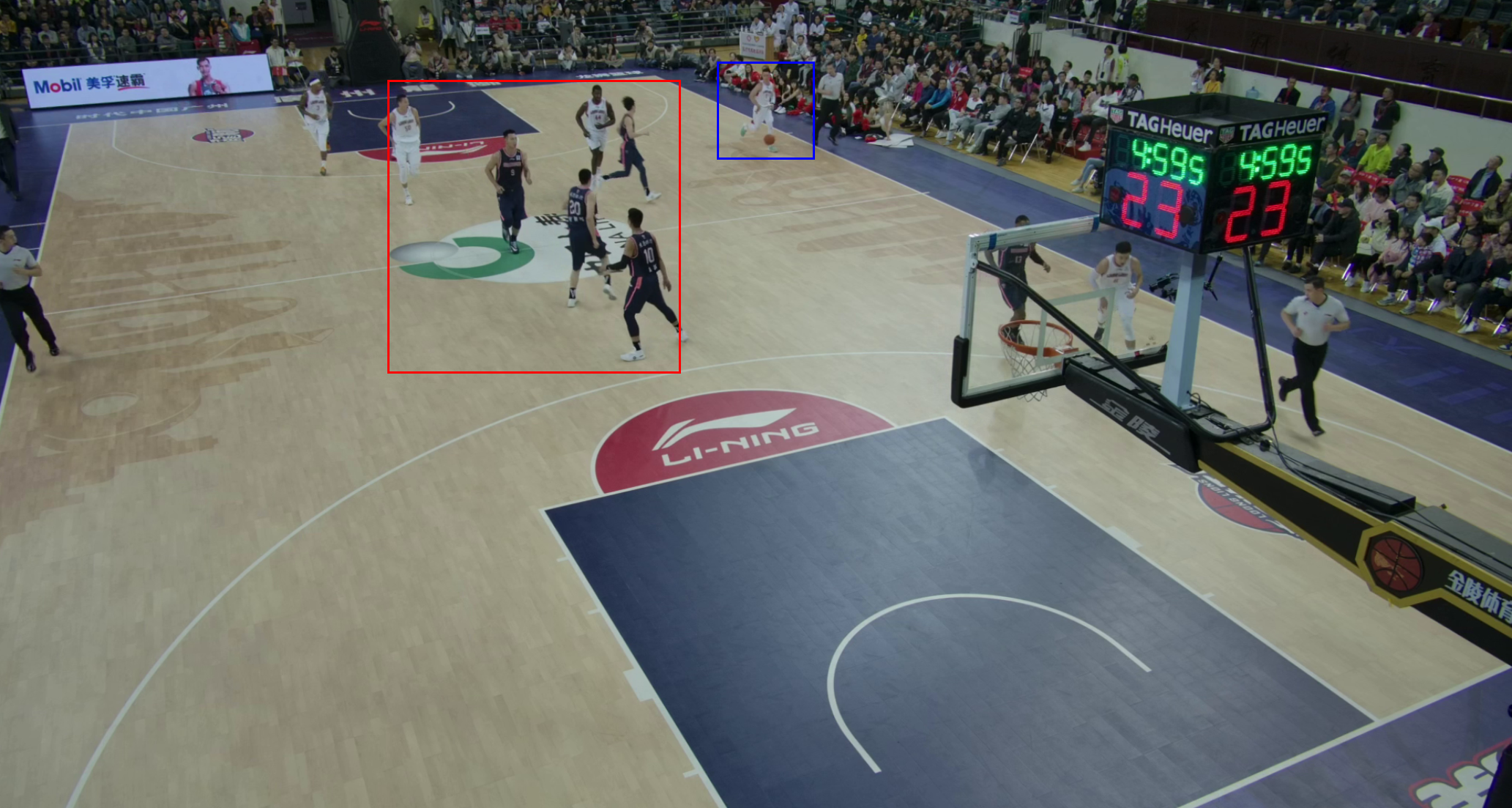} \\
        \includegraphics[width=0.475\linewidth]{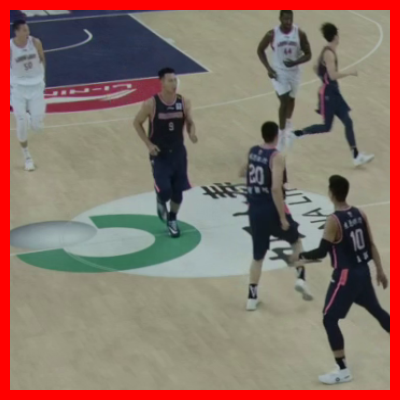}
        \includegraphics[width=0.475\linewidth]{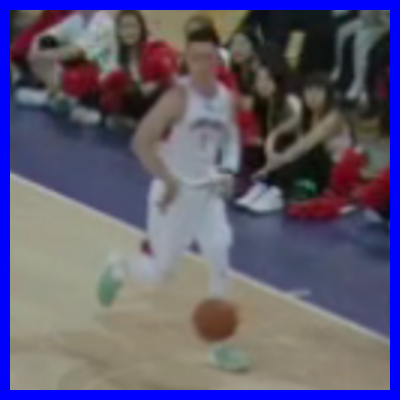}
        \caption*{(a) GT}
    \end{minipage}%
    \hfill
    \begin{minipage}[b]{0.25\textwidth}
        \centering
        \includegraphics[width=0.98\linewidth]{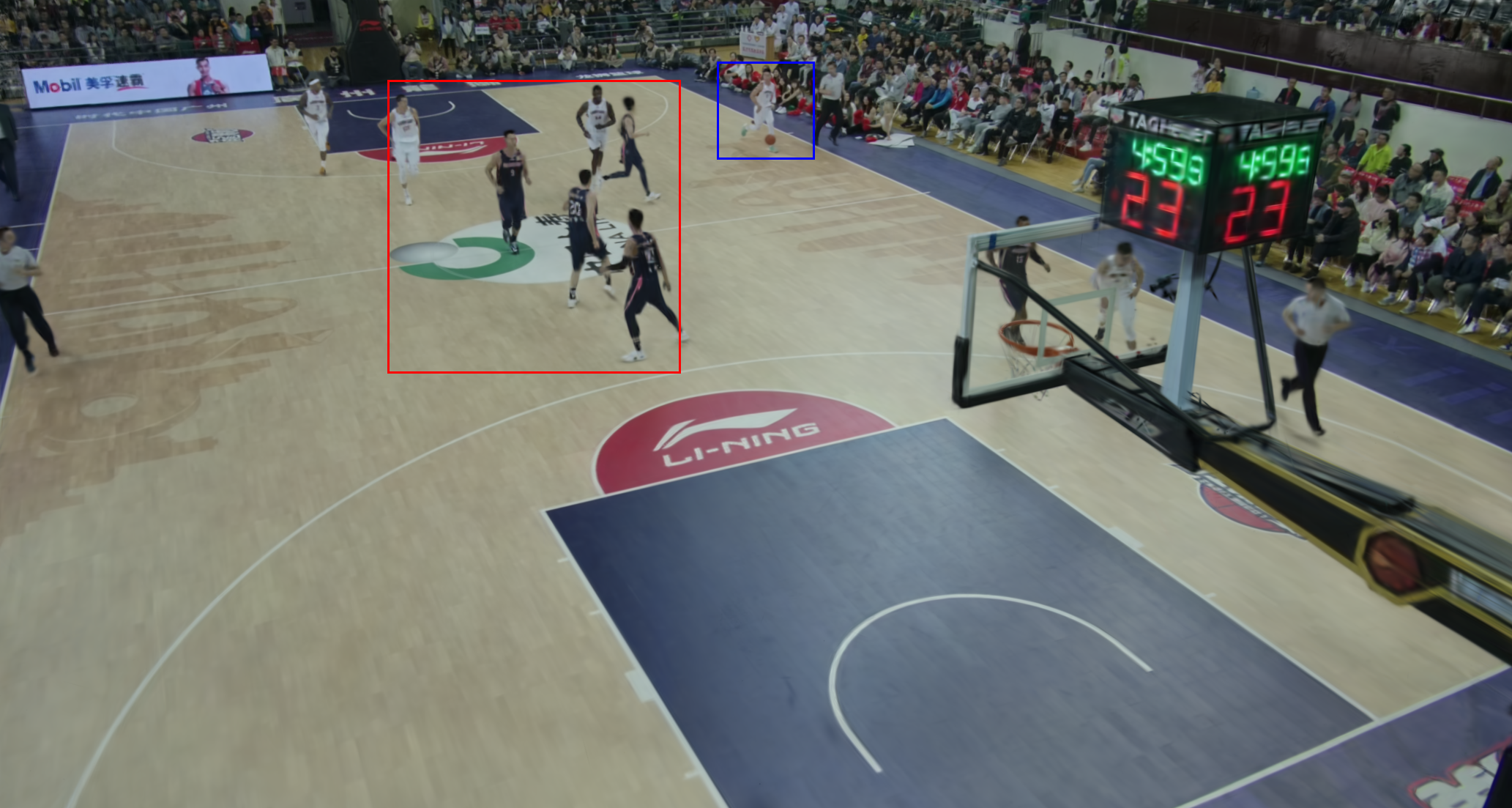} \\
        \includegraphics[width=0.475\linewidth]{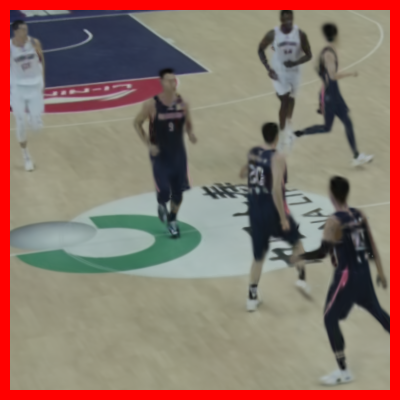}
        \includegraphics[width=0.475\linewidth]{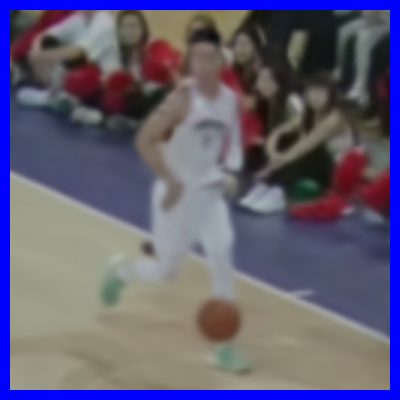}
        \caption*{(b) Ours}
    \end{minipage}%
    \hfill
    \begin{minipage}[b]{0.25\textwidth}
        \centering
        \includegraphics[width=0.98\linewidth]{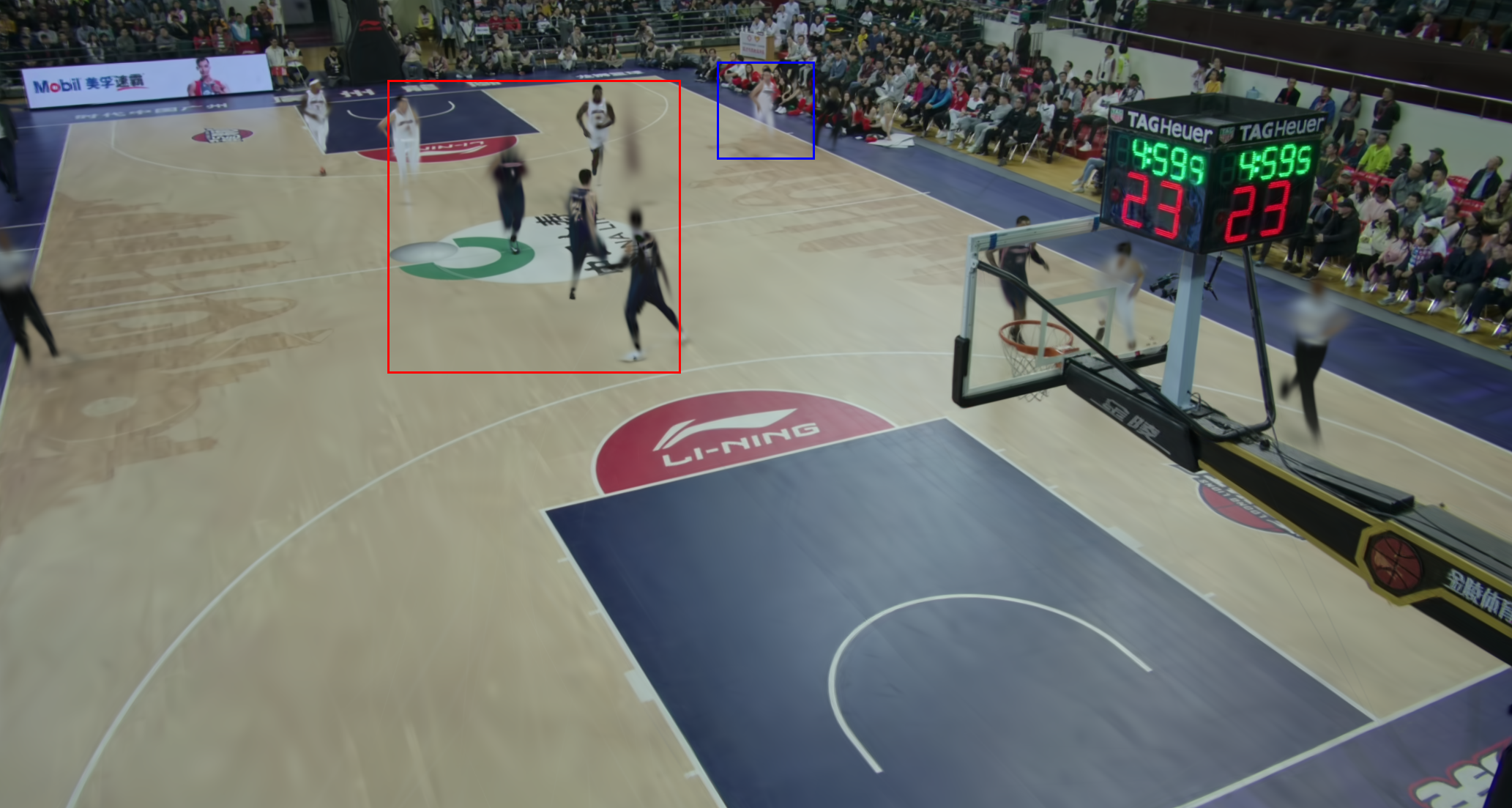} \\
        \includegraphics[width=0.475\linewidth]{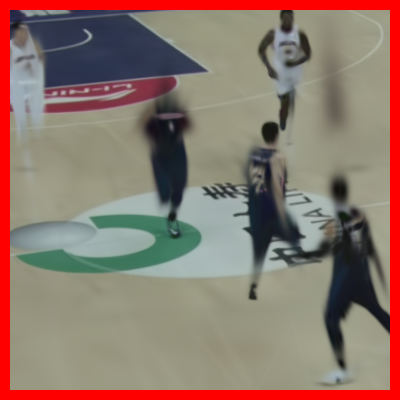}
        \includegraphics[width=0.475\linewidth]{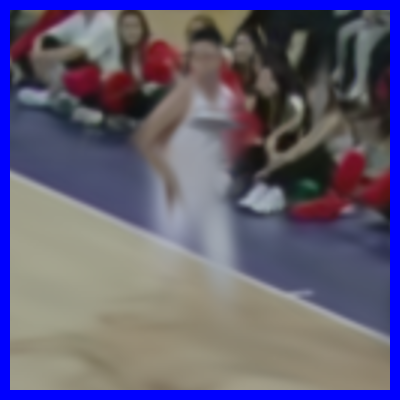}
        \caption*{(c) 4DGS\cite{wu20244d}}
    \end{minipage}%
    \hfill
    \begin{minipage}[b]{0.25\textwidth}
        \centering
        \includegraphics[width=0.98\linewidth]{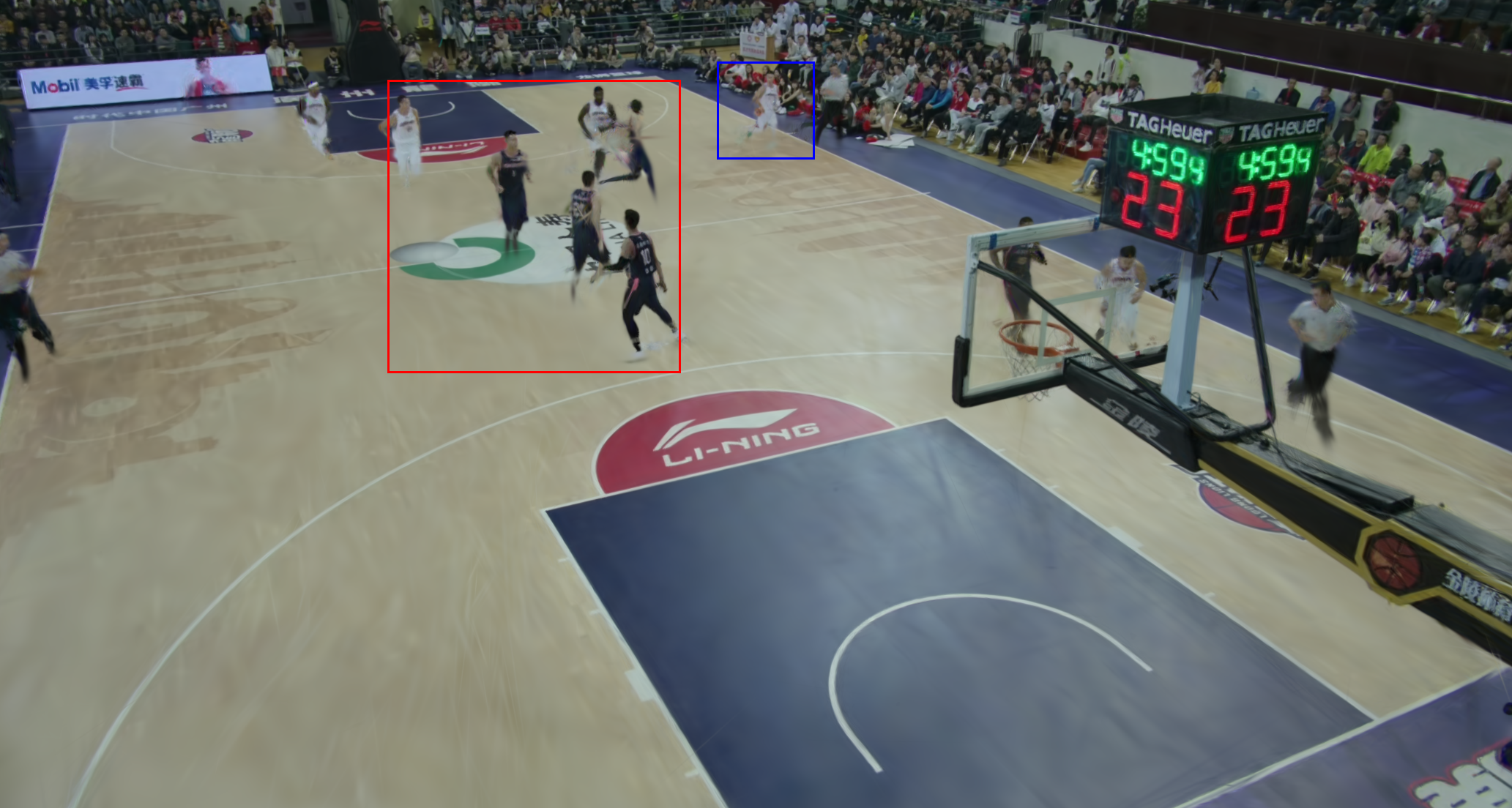} \\
        \includegraphics[width=0.475\linewidth]{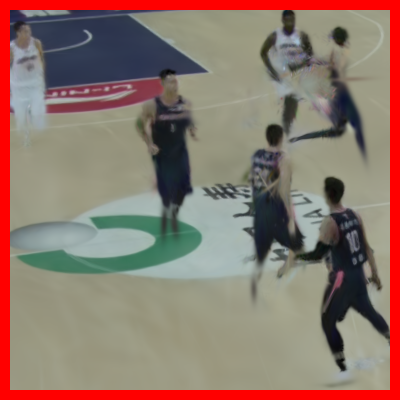}
        \includegraphics[width=0.475\linewidth]{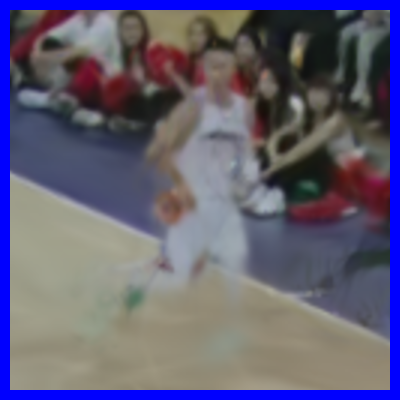}
        \caption*{(d) 3DGStream\cite{sun20243dgstream}}
    \end{minipage}

    \caption{Qualitative results on \textit{VRU GZ} \cite{VRU} (a dataset featuring large-scale, complex motion). Compared to current SOTA dynamic methods, our approach is particularly effective at adapting to large-scale, complex motion scenes. More results can be seen in our videos.}
    \label{fig:vru compare}
\end{figure*}

\subsection{Comparisons}
\label{subsec:Comparisons}
\textbf{Quantitative comparisons.}
We benchmark \ourname~by quantitatively comparing it across the three datasets mentioned above and against a range of SOTA methods, including offline methods like 4DGS \cite{wu20244d} and SpaceTimeGS \cite{li2024spacetime}, as well as the online method 3DGStream \cite{sun20243dgstream}. To verify our method’s outstanding performance, we extract the reported quantitative results on the N3DV dataset from their respective papers, and present the average rendering speed, training time, required storage, PSNR, SSIM, and LPIPS for all scenes in the N3DV dataset in Tab. \ref{tab: n4d}. The results show that our method surpasses previous SOTA methods in multiple aspects, achieving the current SOTA level in quality, while delivering over 10x the speed and requiring only half the storage of the previous SOTA method \cite{li2024spacetime}. 
To demonstrate the generalizability of \ourname, we also conduct experiments on the MeetRoom dataset introduced in StreamRF \cite{li2022streaming}. As shown in Tab. \ref{tab: meetroom all quality comparison}, our method is competitive with the current SOTA streaming method, 3DGStream, particularly excelling in model storage and image quality. Finally, as shown in Tab. \ref{tab: vru compare}, our method demonstrates robust quantitative performance on the VRU basketball dataset \cite{VRU}, which involves larger-scale  motion.

\textbf{Qualitative comparisons.}
We compare scenes from the N3DV dataset and the Meet Room dataset with current mainstream SOTA methods, including the streaming method 3DGStream \cite{sun20243dgstream} and non-streaming methods 4DGS \cite{wu20244d} and SpaceTimeGS \cite{li2024spacetime}. As shown in Fig. \ref{fig:nv3d_comparison}, we particularly highlight the modeling of motion areas in certain scenes, such as hands and claws, as well as complex objects like distant branches and plates. Our method can faithfully capture scene information for both dynamic and complex static objects. Fig. \ref{fig: meetroom compairl} demonstrates the subjective effects in the MeetRoom dataset, where our method outperforms 3DGStream in capturing both dynamic hands and static backgrounds. As shown in Fig. \ref{fig:vru compare}, we also compare our method with the non-streaming 4DGS  and streaming method 3DGStream on the VRU basketball court dataset, which features larger-scale motion. Our approach provides a more faithful representation of large-scale motion.

\begin{table}[t]
    \centering
        \caption{ASG Ablation study on MeetRoom dataset\cite{li2022streaming}. }
    \begin{threeparttable}
    \resizebox{0.8\columnwidth}{!}
    {
        \begin{tabular}{@{}lcccccc}
        \toprule
         Method & PSNR  \(\uparrow\) & $\text{SSIM}_1$\(\downarrow\) & $\text{SSIM}_2$\(\downarrow\) & Size\(\downarrow\) \\
         \midrule
        \multirow{1}{*}{w/o ASG}  &31.81 & 0.021 & 0.011 & 68 MB  \\
        \multirow{1}{*}{w/ ASG }   &33.02 & 0.019 & 0.009 & 96 MB \\
        \bottomrule
        \end{tabular}
    }

\end{threeparttable}
    \label{tab: meetroom asgs ablation}
    \vspace{-1em}
\end{table}

\begin{table}[t]
    \centering
    \caption{\textbf{Ablation study of proposed components.} Conducted on the N3DV dataset~\cite{li2022neural}. }
\begin{threeparttable}
    \resizebox{\columnwidth}{!}
    {
        \begin{tabular}{@{}lccccr@{}}
        \toprule
         Method & \textit{Coffee} & \textit{Sear Steak}   & Mean \(\uparrow\) & FPS  \(\uparrow\) & Time  \(\downarrow\)  \\
         \midrule
        \multirow{1}{*}{w/o static}  & 26.24 & 32.68 & 29.46 & 96 & 42 mins \\
        \multirow{1}{*}{w/o deactivation }   & 29.20 & 33.18 & 31.19 & 89 & 40 mins \\
        \multirow{1}{*}{Full }   & 29.03 & 33.77 & 31.40 & 105 & 36 mins \\
        \bottomrule
        \end{tabular}
    }

\end{threeparttable}
    \label{tab: nv3d ablation}
\end{table}

\begin{table}
\caption{\label{tab:n frames}%
    Ablation study on the number of frames whose SfM point clouds are used in initialization, conducted on N3DV.
}
\renewcommand{\arraystretch}{1}
\centering
\resizebox{0.8\linewidth}{!}{%
\begin{tabular}{@{}lcccr@{\hspace{0.75\tabcolsep}}r@{}}
    \toprule
   N Frames & PSNR$\uparrow$ & DSSIM$_1$$\downarrow$ & LPIPS$\downarrow$ & Time$\downarrow$  \\
  \midrule
  N = 30      & 32.30 & 0.028   & 0.043  & 0.75 h \\
  N = 6      & 32.28 & 0.028 & 0.043   & 0.58 h  \\  
  N = 1     & 31.84 & 0.041  & 0.058   &  0.52 h \\  
\bottomrule
\end{tabular}%
}  
\vspace{-3mm}
\end{table}

\begin{table}[h]
\scriptsize
\centering
\setlength{\tabcolsep}{4pt}  
\caption{Ablation study on different values of $k$ (N3DV dataset).}
\begin{tabular}{lcccccc}
\toprule
\textbf{k value } & PSNR$\uparrow$ & DSSIM$\downarrow$ & Time$\downarrow$ & FPS$\uparrow$ \\
        \midrule
k = 5 & 32.15 & 0.028  & 31.2 m & 118  \\
k = 10  & 32.28  & 0.028 & 34.8 m & 105 \\
k = 20  & 31.96 & 0.032 & 40.5 m & 86 \\
\bottomrule
\end{tabular}
\label{tab: k value}
\vspace{-1em}
\end{table}

\subsection{Ablation Studies}
\label{subsec: ablation}

\begin{figure}[t]
\vspace{-0.3cm}
\centering
\setcounter{subfigure}{0}  
\subfloat[w/o static feature in training]{
		\includegraphics[scale=0.14]{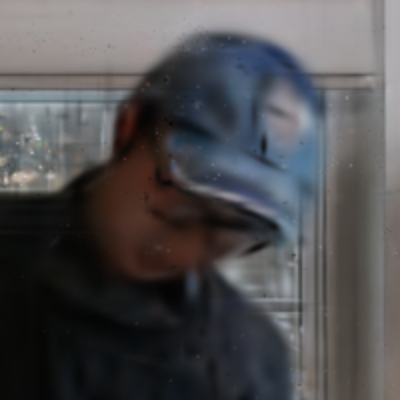}
		\includegraphics[scale=0.14]{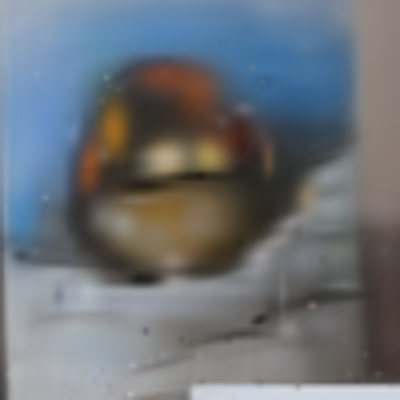}}
\subfloat[Full]{
		\includegraphics[scale=0.14]{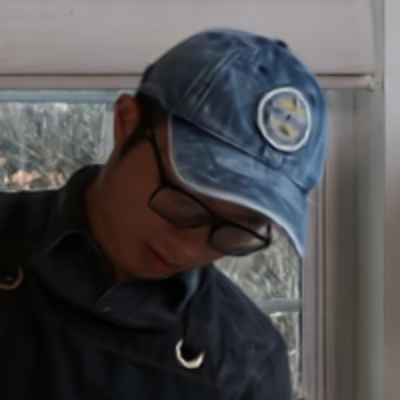}
		\includegraphics[scale=0.14]{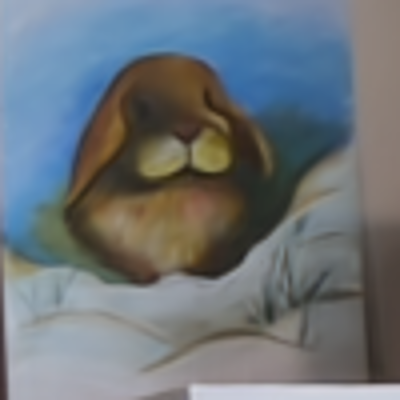}}
\caption{ A comparison of (a) to (b) shows that training using only dynamic features leads to significant blurring issues.}
\label{fig: wo static}
\end{figure}

\textbf{Decoupling of dynamic-static feature.} To validate our static-dynamic feature decoupling approach, we remove the static feature and retain only the dynamic residual feature for training. As shown in Fig. \ref{fig: wo static} and Tab. \ref{tab: nv3d ablation}, removing the static feature leads to a noticeable decline in rendering quality. We hypothesize that this limitation arises from the dynamic residual field's inability to encode the full scene information (both static and dynamic), leading to noticeable blurring and distortion effects.

\begin{figure}[t]
    \centering
    
\vspace{-0.4cm}
    \begin{minipage}[b]{0.235\textwidth}
        \centering
        \includegraphics[width=0.97\linewidth]{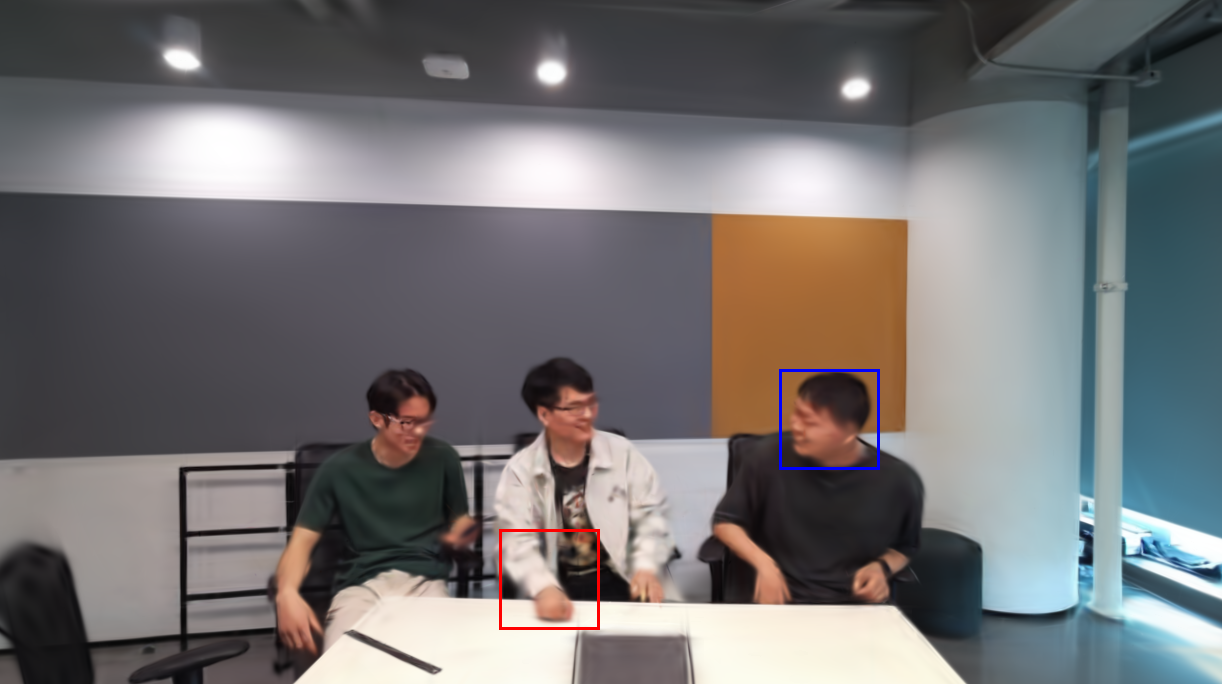} \\
        \includegraphics[width=0.475\linewidth]{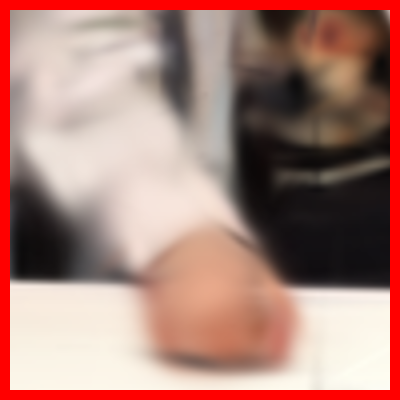}
        \includegraphics[width=0.475\linewidth]{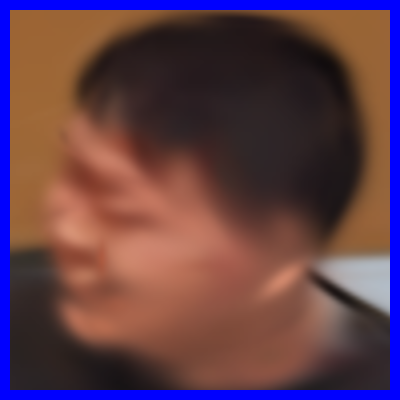}
        \caption*{(a) w/o ASG}
    \end{minipage}%
    \hfill
    \begin{minipage}[b]{0.235\textwidth}
        \centering
        \includegraphics[width=0.97\linewidth]{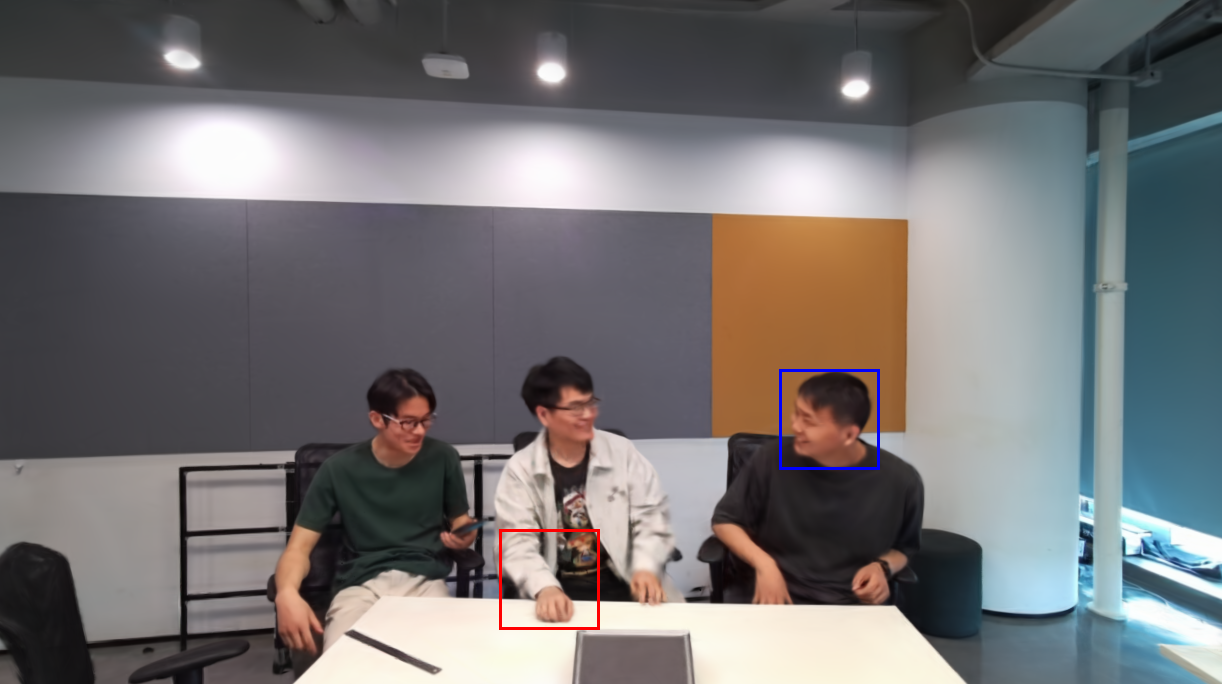} \\
        \includegraphics[width=0.475\linewidth]{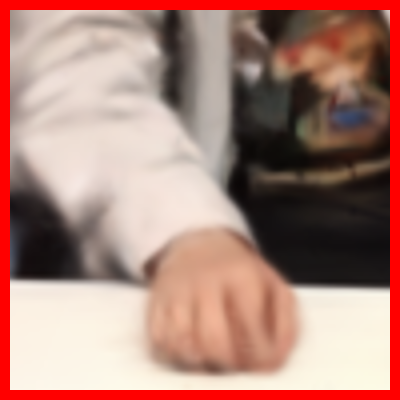}
        \includegraphics[width=0.475\linewidth]{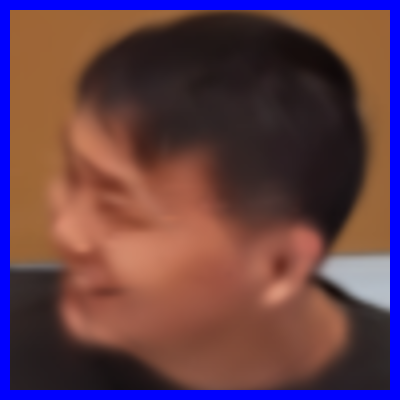}
        \caption*{(b) w/ ASG}
    \end{minipage}
    
    \caption{Ablation study conducted on the \textit{discussion} scene.}
\label{fig: asgs ablation}
    \vspace{-1em}
\end{figure}

\textbf{Adaptive seed growing (ASG).} We conduct ablation experiments on our ASG (Sec. \ref{subsec: asgs}) in the \textit{discussion} scene. As shown in Fig. \ref{fig: asgs ablation} and Tab. \ref{tab: meetroom asgs ablation}, the ASG technique demonstrates a clear improvement in the accuracy of dynamic region reconstruction.

\textbf{The number of frames used for initialization.} As shown in Tab. \ref{tab:n frames}, the performance of our method increases with the number of initialization frames. In the experiments, we set \(N=6\) to balance performance and training time.

\begin{figure}
   \centering
   \includegraphics[width=0.45\textwidth, trim=0cm 10cm 11cm 0cm]{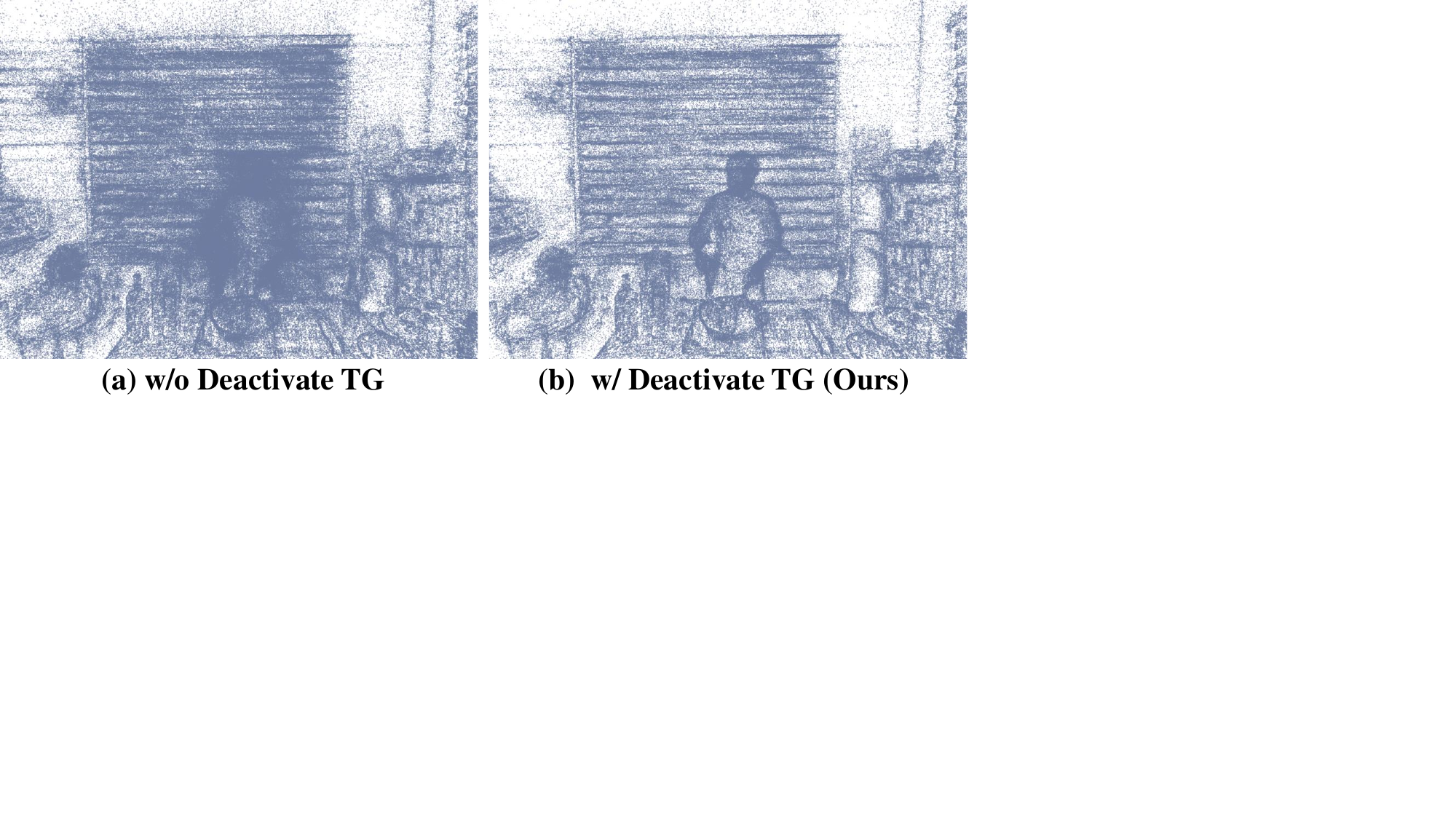}
   \caption{Our deactivation strategy effectively reduces a significant amount of redundant Temporal Gaussians.}
   \label{fig: inactive ng}
   \vspace{-10pt}
\end{figure}

\textbf{The deactivation of temporal Gaussians.} As shown in Fig. \ref{fig: inactive ng} and Tab. \ref{tab: nv3d ablation}, this approach effectively reduces redundant Temporal Gaussians without compromising rendering quality, leading to a substantial improvement in inference speed.

\textbf{Learning with different $k$ value.}  We apply different $k$ values in our method as same as ScaffoldGS \cite{lu2024scaffold}. The results as shown in Tab. \ref{tab: k value}.

\section{Limitation and Discussion}
\label{sec:discussion}

Similarly to previous methods \cite{li2024spacetime, sun20243dgstream}, our approach relies on point clouds estimated by SfM for initialization. If SfM fails significantly, it may impact rendering quality. However, we have not encountered such a case so far, even when SfM failed in the MeetRoom dataset. Meanwhile, since point cloud estimation and densification are orthogonal to dynamic reconstruction, we have not focused on this issue in detail.
In addition, our task uses multi-view synchronized video for dense spatiotemporal supervision to output high-quality dynamic videos from free viewpoints. We also believe that a pre-trained model providing complete geometric information for monocular input could enable high-quality dynamic video construction from monocular input.

\section{Conclusion}
\label{sec:conclusion}

In this paper, we first introduce a method that can adapt not only to fine-scale dynamic scenes but also to large-scale scenes.
Specifically, we propose decomposing the 3D space into local space based on seeds. For motion modeling within each local space, we assign a static feature shared across all time steps to represent static information, while a global dynamic residual field provides time-specific dynamic residual features to capture dynamic information at each time step. Finally, these features are combined and decoded to produce time-varying Temporal Gaussians, which serve as the final rendering primitives.
Extensive experiments show that \ourname~effectively adapts to dynamic scenes across various motion scales, performing well on both large-scale motion datasets, such as the basketball court \cite{VRU}, and fine-scale motion datasets, like N3DV \cite{li2022neural} and MeetRoom \cite{li2022streaming}.
We hope the proposed local motion modeling approach offers new insights for dynamic 3D scene modeling.

{
    \small
    \bibliographystyle{ieeenat_fullname}
    \bibliography{main}
}

\end{document}